# Hybrid Firefly-Genetic Algorithm for Single and Multi-dimensional 0-1 Knapsack Problems

**Aswathi Malanthara and Ishaan R Kale***

Institute of Artificial Intelligence, Dr Vishwanath Karad MIT World Peace University

Pune: 411038, Maharashtra, INDIA

1032210470@mitwpu.edu.in; *ishaan.kale@mitwpu.edu.in

## Abstract

This paper addresses the challenges faced by algorithms, such as the Firefly Algorithm (FA) and the Genetic Algorithm (GA), in constrained optimization problems. While both algorithms perform well for unconstrained problems, their effectiveness diminishes when constraints are introduced due to limitations in exploration, exploitation, and constraint handling. To overcome these challenges, a hybrid FAGA algorithm is proposed, combining the strengths of both algorithms. The hybrid algorithm is validated by solving unconstrained benchmark functions and constrained optimization problems, including design engineering problems and combinatorial problems such as the 0-1 Knapsack Problem. The proposed algorithm delivers improved solution accuracy and computational efficiency compared to conventional optimization algorithm. This paper outlines the development and structure of the hybrid algorithm and demonstrates its effectiveness in handling complex optimization problems.

## 1. Introduction

Nature-inspired optimization methods have become crucial techniques for solving complex optimization problems (Cheng et al., 2014) by leveraging principles from natural and biological systems. Heuristics approaches (Newell et al., 1958) use problem-specific, rule-based strategies. Whereas, metaheuristics approaches (Glover and Kochenberger, 2003) provide more flexible frameworks that adapt to various optimization tasks. Algorithms like Genetic Algorithm (GA) (Holland, 1975; Bernardino et al., 2007), which mimics natural phenomenon through selection, crossover, and mutation, are known for maintaining diversity, although might face challenges with longer time for convergence while not ensuring to reach the optimal solution. Particle Swarm Optimization (PSO) (Kennedy and Eberhart 1995; Datta and Figueira, 2011), based on the behavior of bird flocks, is known for fast convergence. However, it may become trapped in local optima if exploration is insufficient. Other algorithm includes, Ant Colony Optimization (ACO) (Dorigo et al., 1996), which simulates ant foraging and is particularly effective for discrete problems, however may become costly with larger datasets. Differential Evolution (DE) (Storn and Price, 1995; Karaboğa and Ökdem, 2004) models the process of evolutionary adaptation and is effective for continuous optimization, yet may require precise parameter adjustments. The Bat Algorithm (BA) (Yang 2010; Chakri et al., 2017), inspired by bat echolocation, offers versatility yet encounter difficulties in tuning its randomness. Harmony Search (HS) (Geem et al., 2001; Yang, 2009), draws from the concept of musical improvisation, is simple and effective although exhibit slow convergence on more complex problems. Cuckoo Search (CS) (Yang and Deb, 2009), which imitates the parasitic nesting or egg parasitism behavior of cuckoos, excels at global optimization, even so it struggles with convergence speed when constraints are high. The Firefly Algorithm (FA) (Yang, 2010; Gandomi et al., 2011), based on firefly bioluminescence, performs well in multimodal scenarios while face challenges like early convergence, balancing exploration and refinement. Simulated Annealing (SA) (Kirkpatrick et al, 1983), reflecting the physical process of annealing metals, is easy to implement though might be slow in high-dimensional spaces. These algorithms share common drawbacks, such as significant computational cost, parameter sensitivity, and potential for becoming trapped in local solution. They effectively manage non-linear and multi-modal problems. However, their unconstrained nature may limit their efficiency in constrained or combinatorial problems (Blum and Roli, 2003).

Hybrid algorithms in optimization integrate multiple algorithms to enhance performance by combining their strengths. The Genetic Algorithm-Tabu Search (GA-TS) hybrid algorithm (Lin et al., 2010) illustrates this integration by leveraging GA's broad search capabilities, including selection, crossover, and mutation, while Tabu Search (TS) performs local searches and uses a memory structure to avoid revisiting past solutions, promoting a balance between exploration and exploitation. Hybrid genetic algorithm and simulated annealing approach (GA-SA) (Li et al., 2002) explores a wide search space using GA, with simulated annealing refining solutions to avoid local optima. The PSO-GA algorithm (Garg, 2016) combines PSO's rapid global search, driven by particle movement and position updates, with GA's genetic operations to enhance optimization. Genetic algorithm - differential evolution (GA-DE) (Lin, 2011) integrates GA's exploration capabilities with DE's adaptive parameter control, achieving robust performance in global optimization. Despite these strengths, each hybrid algorithm has limitations: GA-TS faces challenges with high computational costs and complex coordination between algorithms; GA-SA demands significant computational resources due to the combination of two iterative methods; PSO-GA struggles with balancing exploration and exploitation, occasionally leading to premature convergence; and GA-DE faces challenges in parameter tuning, which impacts convergence speed and solution accuracy in some applications. The Ant Colony Optimization-Genetic Algorithm (ACO-GA) (Ghanbari et al., 2013) blends ACO's pheromone-based navigation with GA's genetic diversity from crossover operations, enhancing exploration and reducing early convergence risk. This approach is similar to Firefly and Particle swarm optimization (HFPSO) (Aydilek, 2018), which combines the global search ability of the FA with the local search and optimization strengths of PSO. The FA in HFPSO attracts particles toward brighter solutions, ensuring effective global search, while PSO guides particles based on their individual and collective experiences to refine the search locally. However, HFPSO lies in its computational cost, especially for large-scale or high-dimensional optimization problems, and its sensitivity to initial conditions due to the complex interaction between FA and PSO parameters, which impacts the balance between global exploration and local exploitation. The Hybrid Firefly algorithm and Harmony search (HS-FA) (Guo et al., 2013) combines the global exploration strength of the Firefly Algorithm with the local search power of Harmony Search, where FA explores the search space by moving toward brighter solutions, and HS refines these solutions by adjusting harmony memory and applying pitch adjustment operations. Although, HS-FA face the challenge of fine-tuning the harmony memory size and pitch adjustment parameters, leading to computational demands that increase with problem size and slow convergence in complex search spaces. The Imperialist competitive algorithm and Firefly algorithm (ICA-FA) (Chen et al., 2018) integrates the exploration power of the ICA with the FA search capabilities. ICA directs the search towards high-potential regions through imperialistic competition, while the FA intensifies the search around the best solutions by attracting to brighter solutions. ICA-FA lies in its reliance on imperialist competition, which results in premature convergence, particularly in multi-modal problems, with parameter dependencies complicating its implementation and fine-tuning. The Firefly Algorithm-Cuckoo Search (FA-CS) (Yang, 2014) hybrid combines the global random exploration of Cuckoo Search using Levy flights with the Firefly Algorithm's local search ability to enhance exploration while refining solutions in promising regions. However, the combination of both algorithms results in increased computational complexity and requires extensive parameter tuning. Overall, Hybrid algorithms improve performance over single-method algorithms, but they introduce challenges such as higher computational costs, parameter sensitivity, and complex implementation. Achieving optimal performance requires careful parameter tuning and balancing the strengths and weaknesses of the combined methods.

This paper introduces a hybrid algorithm that combines the FA (Yang, 2010; Gandomi et al., 2011) and the GA (Holland, 1975; Bernardino et al., 2007) to address complex real-world problems, such as the 0-1 Knapsack Problem, benchmark functions, and engineering design problems. An earlier approach (Nand and Sharma, 2019) segmented the optimization process into two phases: first, FA is used for global exploration in the first half, followed by GA for local refinement in the second half. While this approach achieved reasonable success in balancing global and local search strategies, it also had limitations, such as slower convergence rates and reduced solution diversity. These issues arose because FA and GA are executed independently in different stages, limiting their ability to fully complement each other. By operating in isolation, the algorithms could not leverage their respective strengths effectively throughout the entire optimization process. In contrast, the proposed hybrid algorithm integrates FA and GA in a simultaneous, continuous manner, allowing both to work together throughout the optimization process. This concurrent framework addresses the limitations of the earlier method by enabling FA to continuously refines the population. Whereas, GA enhance the population's exploration through selection, crossover, and mutation. This dynamic interaction helps maintain diversity and prevents premature convergence. The significant enhancement of the proposed algorithm is the seamless integration of FA's bioluminescence inspired movement with GA's genetic operations. Through which it guides the algorithm towards better diversity and more efficient solution refinement. This integration enables a balance exploration and exploitation over the optimization process, resulting in faster convergence and higher-quality solutions compare to previous methods. To evaluate the performance of the proposed algorithm, it is tested on a wide range of optimization problems. These include the benchmark optimization functions (Zhang et al., 2016; Qi et al., 2017), such as the Sphere function (unimodal function), the Ackley function, Rosenbrock's function, and the Rastrigin function (multimodal functions). These unconstrained functions are commonly used to test the performance of optimization algorithms. Additionally, the FAGA hybrid has been applied to real-world engineering design optimization problems (Kale and Kulkarni, 2018; Kale and Kulkarni, 2021), including the helical spring design problem, pressure vessel optimization, cantilever beam optimization, gear train ratio, and I-beam vertical deflection. These problems involve both continuous and discrete variables, making them highly suitable for testing the versatility and robustness of the proposed algorithm. Furthermore, the 0-1 Knapsack Problems (Beasley and John, 1990; Kulkarni and Shabir, 2016; Poonawala et al., 2024), examples of combinatorial optimization, are also addressed. The SKP (Single Knapsack Problem) involves selecting a subset of items to maximize the total value without exceeding a given weight limit, while the MKP (Multidimensional Knapsack Problem) introduces additional constraints by increasing the number of knapsacks, making the problem significantly more complex. These types of problems are crucial in areas like cargo loading, financial investment planning, and resource allocation, where multiple constraints must be satisfied simultaneously. The results demonstrate that the proposed FAGA hybrid consistently outperforms, matches, or achieves near-optimal performance, both in terms of solution quality and convergence speed.

The paper is organized as follows: Sect. 2 introduces the basic principles and mathematical formulation of the FA, along with its characteristics and pseudo code. In Sect. 3, described GA's evolutionary operators, pseudo code, and characteristic ability to maintain diversity during the search process. The hybrid FAGA algorithm is proposed in Sect. 4, where the integration of FA and GA is explained along with its flowchart. In Sect. 5, explanation about integration of static penalty function to handle constraint violations. In Sect. 6, the significance and formulation of the 0-1 Knapsack Problem are presented, focusing on maximizing profit while staying within weight limits. The methodology employed by FAGA to solve the 0-1 Knapsack problem is detailed, showing how FA's exploration is combined with GA's crossover and mutation strategies, along with a flowchart In Sect. 7, various optimization problems are solved, including benchmark functions, design engineering problems, and the 0-1 single and multidimensional knapsack problems using the FAGA algorithm. A statistical analysis of the FAGA algorithm, along with a comparison to other algorithms, is presented to evaluate its performance, accompanied by its convergence graphs. The list of 0-1 SKP test cases is provided in the "Appendix" at the end of the paper.

## 2. Firefly Algorithm (FA)

The Firefly Algorithm (Yang, 2010; Gandomi et al., 2011), introduced by Xin-She Yang , is a nature-inspired metaheuristic that draws on the principles of swarm intelligence. The algorithm is influenced by the behavior of fireflies, using randomization to search for a set of solutions, and is thus classified as a stochastic algorithm. The flashing behavior of fireflies is modeled as a mechanism to attract prey or mates. The fitness or intensity represents the brightness of a solution; a firefly is considered brighter when it has a better solution compared to others. Less bright fireflies move toward the brighter ones, helping them move in the direction of a better solution in the search space.

Additionally, distance plays a major role, as the attractiveness between fireflies decreases with increasing distance. This is crucial in determining the level of attraction one firefly has towards another. The pseudo code of the FA is presented in Fig. 1. Following are the characteristics of Firefly Algorithm:

(1) Fireflies are drawn to each other without regard to gender, as they are unisex.
(2) The level of attractiveness increases with the intensity of brightness.
(3) Fireflies will move toward the brighter ones, however if no other firefly is brighter or if brightness levels are equal, they will move randomly.
(4) Attractiveness diminishes with increasing distance, meaning it is inversely related to the distance between fireflies.

The FA is mathematically expressed as follows:

***Step 1:*** Consider a population consisting of $n$ fireflies, where each firefly $x_i$ $(for\ i = 1, 2, \dots, n)$ is defined by a vector of decision variables representing its position in the search space. Initially, the positions of the fireflies are generated randomly within specified lower $(lb)$ and upper $(ub)$ bounds. The initial position of each firefly is given by:

$$X^n = lb + (ub - lb) \times rand(1, M), \quad (2.1)$$

where, $M$ denotes the number of decision variables for each firefly. This equation ensures that each firefly starts at a valid location in the search space, effectively initializing their positions based on the specified boundaries.

***Step 2:*** After initializing the $n$ fireflies, the $f(X_n)$ for each firefly is calculated. The algorithm then determines the current best fitness value among all the fireflies in the population.

***Step 3:*** Each firefly $i$ is compared with every other firefly $j$ in the population based on intensity. If firefly $i$ has a higher intensity than firefly $j$, no change occurs, and it moves to the next firefly However, if firefly $j$ has a greater intensity, indicating a better solution, firefly $i$ moves toward firefly $j$. The movement of firefly $i$ towards firefly $j$ is influenced by three key factors: attractiveness, randomness, and distance. These factors combine to guide the firefly's movement and position update. The movement equation is formulated as follows:

$$Xinew = Xi + \beta_0 e^{-\gamma r_{ij}^2}.(Xj - Xi) + \alpha(rand - 0.5), \quad (2.2)$$

where,

$\beta_0$ is the highest possible attraction when the distance between fireflies is zero,

$\alpha$ is the randomization parameter, regulating the amount of randomness in the movement
$\gamma$ is the light absorption coefficient, which controls how quickly attractiveness diminishes with distance,

$$r_{ij} = |X_i - X_j| = \sqrt{\Sigma_{k=1}^{M}(x_{ik} - x_{jk})^2}, \quad (2.3)$$

is the Euclidean distance between firefly $i$ and firefly $j$,
$rand$ is a random vector with values between 0 and 1,
$X_i$ and $X_j$ are the current positions of firefly $i$ and firefly $j$, respectively.
This equation updates the position of firefly $i$ by moving it towards the brighter firefly $j$, while the random component helps to avoid being trapped in local minima.

***Step 4:*** The algorithm halts if there has been no improvement in the best solution over a defined number of iterations. The termination condition is also based on reaching the maximum number of iterations, $max_iter$.

```
α            Randomization parameter
β0           Attractiveness constant
γ            Light absorption coefficient
n            Number of fireflies
max_iter     Maximum number of iterations

Initialize α , β0 , γ , n , max_iter
    Generate an initial population of n for fireflies x_i ( i = 1,2, … , n).
    Evaluate Fitness for individual firefly /Calculate objective function f(x_i)
    While ( t < max_iter),
        For i = 1 to n
            For j = 1 to n
                If (f(x_j) > f(x_i))
                    Move firefly i towards j ;
                    Update firefly i's position:
                      x_i(new) = x_i(old) + β0 * exp ( − γ * rij ^2) * (xj − xi) +
    α (rand() − 0.5)
                End if
              Evaluate fitness value of new x_i
            End for j
        End for i
        Rank the fireflies and find the current best solution
End while
4.Postprocess results and visualization
```

**Fig. 1.** Pseudo code of FA

The FA (Yang, 2010; Gandomi et al., 2011) is successfully validated on various continuous, discrete, and mixed-variable optimization problems. It had been widely applied in design engineering problems, multi-modal optimization, and structural optimization problems. It is observed that the convergence rate of the FA varies depending on the complexity of the problem and the parameters chosen, such as attractiveness, randomness and absorption coefficient. In the standard FA, the brightest firefly moves randomly, which may lead to a decrease in brightness depending on the direction. As a result, the performance of the algorithm diminishes. Furthermore, researchers had proposed various enhancements to the FA, leading to better performance and faster convergence. One such modification is the Modified Firefly Algorithm (MFA) (Yelghi et al., 2018), which integrates adaptive mechanisms where the brightest firefly moves only in a direction that improves its brightness. This helps fireflies explore the search space efficiently and avoids premature convergence. The MFA has been successfully validated by solving test problems that include multimodal, unimodal, stochastic, continuous and discontinuous benchmark problems.

## 3. Genetic Algorithm (GA)

The Genetic Algorithm (Holland, 1975; Bernardino et al., 2007), proposed by John Holland, is an adaptive metaheuristic algorithm inspired by the principles of natural selection. In this algorithm, Selection process is typically probabilistic, favoring individuals with higher fitness. Further, Selected individual undergo mating, where recombination of the genetic information of two parents creates new, better offspring that inherits beneficial characteristics from the parents. The child having the certain characteristics of both parents, is mutate to introduce variability and diversity, which help to enhance a broader exploration of the search space, ensuring to reach optimal solution across a range of possibilities. The pseudo code of the GA is presented in Fig. 2. Following are the characteristics of GA:

1) Genetic Algorithms operate on group of potential solutions rather than single solutions while simultaneously exploring multiple points in the search space.
2) Individuals are evaluated using a fitness function, and the best performing solutions are selected for reproduction.
3) Crossover combines genes from two parents to create child, while mutation integrate random changes. This ensures diversity and exploration.

The GA is mathematically expressed as follows:

***Step 1:*** An initial population of $n$ individuals, denoted as $x_i$ $(for\ i = 1, 2, \dots, n)$, is generated randomly within the defined search space. Each individual is represented by a set of decision variables, and their initial positions are established using the same formula as shown in Equation (2.1).

***Step 2:*** The fitness or objective function $f(X_n)$ is computed for each individual in the population to evaluate their performance in the optimization process.

***Step 3:*** Parents are selected using the *Tournament Selection* method. A predefined number of individuals, referred to as the tournament size $T$, are randomly chosen from the population. The individual with the highest fitness within this group is selected as a parent. It is mathematically expressed as:

$$p_{winner} = arg_{i \epsilon Tournament} \max f(x_i), \tag{3.1}$$

Where $p_{winner}$ represents the chosen parent, Tournament is the set of randomly selected individuals, and $f(x_i)$ is the fitness of individual $i$. This process is repeated until the required number of parents is selected for crossover.

***Step 4:*** Crossover is performed based on the defined crossover rate. Two parents, $p_1$ and $p_2$, are chosen to create offspring. The children are generated using crossover points, which is mathematically represented as:

$$c = \alpha . p_1 + (1 - \alpha). p_2, \tag{3.2}$$

where $\alpha$ is a mixing parameter ranging from 0 to 1. If the offspring's fitness is superior to that of its parents, the child is kept; otherwise, the parents are retained for the next generation.

***Step 5:*** A mutation process is conducted according to the specified mutation rate. Individuals are randomly selected, and a mutation operator is applied to introduce variation into the population. For Gaussian mutation, this is be expressed as:

$$x_i' = x_i + N(0, \sigma), \tag{3.3}$$

Where $N(0, \sigma)$ denotes a random value sampled from a normal distribution with mean 0 and standard deviation $\sigma$, influencing the mutation strength.

***Step 6:*** The fitness of the new population is compared to that of the previous population. If the new fitness values show improvement, the new population is accepted; if not, the original population is retained.

***Step 7:*** If no significant changes are observed, it suggests that the solution has stabilized, and the fitness of the original population is accepted, resulting in termination of the algorithm. If changes continue to be evident, the process loops back to ***Step 2*** for further iterations.

GA (Holland, 1975; Bernardino et al., 2007) had proven to be an effective method for solving real-world optimization challenges. GA maintains population diversity exploration to prevent the loss of relevant information, resulting in a balanced search. GA performs well as a global search method; it may take time to converge and may not guarantee to reach optimal solution. Further, GA was enhanced by integrating it with two gradient descent (GD)-based algorithms (Ruder, 2016), leading to the development of the Gradient-based Genetic Algorithm (GGA) (D'Angelo et al., 2021). This algorithm has the capability to identify the optimal solution with fewer generations and individuals. The basic idea involves leveraging GD's capabilities to refine local solutions and employing them as more favorable starting points. This approach enables GGA to escape local optima and progressively converge toward the global solution. The GGA has been successfully validated by solving test functions characterized by multi-modality, flatness, and convexity, as well as in addressing a real-world use case: the welded beam design problem.

```
    cross_rate     Crossover rate
    mut_rate       Mutation rate
    n              Number of populations
    max_iter       Maximum number of iterations

Initialize α, cross_rate , mut_rate, n, max_iter
  1. Generate an initial population of n individuals, x_i (i = 1, 2, ..., n).
  2. Evaluate the fitness/objective function f(x) for each individual.
  3. Select parents based on fitness.
  4. Apply crossover with crossover rate cross_rate:
       - Select two parents, p_1 and p_2.
       - Generate children using crossover points.
          If (child's fitness is better than its parent's)
             return child
          Else
             return parents
          End If
  5. Apply mutation with mutation rate mut_rate:
        - Select candidates randomly and apply mutation operator.
  6. If (new_fitness better than old_fitness)
        - Return the new population.
     Else
        - Return the original population.
    End If
  7. Update the population n with the new population and fitness.
  8. If: there is no significant change in fitness, consider the solution saturated:
       Accept fitness of the original population and terminate.
    Else
      Continue to Step 2.
    End if
End While
9. Postprocess results and visualization.
```

**Fig. 2.** Pseudo code for GA

## 4. Framework of FAGA

The standard Firefly Algorithm (FA) has been validated across a diverse array of optimization problems. However, it faces several limitations, such as an imbalance between exploration and exploitation, reduced local convergence when the randomization factor is high, and a tendency to fail in finding the optimal solution due to limited local and global search capabilities. To address these issues, key characteristics of the Genetic Algorithm (GA) are integrated into the FA. The GA helps to balance exploration and exploitation by generating diverse solutions. Essential GA operators - Selection, Crossover, and Mutation are incorporated into the FA to enhance local search and convergence. In the FAGA framework (refer to Fig. 3), the FA primarily facilitates global search, where fireflies are attracted to those with higher intensities (better solutions), while the GA introduces genetic diversity through crossover and mutation. This integration enhances the learning ability of the FA. The FAGA is mathematically outlined as follows:

***Step 1:*** Consider a population of $N$ individuals, where each individual $n$ $(for\ n = 1, 2, \dots, N)$ is characterized by a set of decision variables $X^n = (x_1^n, x_2^n, \dots, x_M^n)$, representing the individual's position in the search space. The intensity of each individual is calculated based on an objective function $f(X^n)$. The initial position of each individual is randomly generated within the specified lower $(lb)$ and upper $(ub)$ bounds, using the same approach as shown in equation $(2.1)$.

***Step 2:*** Each firefly is compared with every other firefly in the population. If the intensity of firefly $i$ is better than that of firefly $j$, no movement occurs, and firefly $i$ proceeds to next firefly. However, if firefly $j$ has a higher intensity (better solution) than firefly $i$, firefly $i$ will move towards firefly $j$. The movement of firefly $i$ towards firefly $j$ is influenced by three key factors: attractiveness, randomness, and distance. The movement equation is formulated as shown in equation $(2.2)$.

***Step 3:*** Once all comparisons are complete, the intensities $f(X^n)$ of the fireflies updated positions are calculated, resulting in the generation of a new population of fireflies.

***Step 4:*** After generating the new population and calculating their fitness, the best-performing individuals are selected using tournament selection, as described in equation $(3.1)$, for crossover. The crossover rate is maintained between *60%* and *90%* to ensure diversity. For two selected parents, $p_1$ and $p_2$ , crossover is executed by combining their characteristics to create a child $c$. This crossover operation is mathematically represented in equation $(3.2)$. After crossover, the fitness $f(X^n)$ of each child is evaluated. If the child's intensity surpasses that of its parents, the child proceeds to mutation; otherwise, the original individuals are retained.

***Step 5:*** Following crossover, a mutation rate of *1-10%* is applied to mutate individuals. During mutation, some attributes are altered to enhance diversity in the population. Gaussian mutation is employed, as represented in equation $(3.3)$. The mutated child then replaces the worst-performing individual in the population. The intensity of the new individual is calculated, and the top solutions are carried over to the next iteration.

***Step 6:*** The algorithm terminates when there is no further change in the solution, indicating stagnation, or when the termination criteria (number of iterations) are met.

To validate the proposed FAGA algorithm, the problems considered here are sourced from the design engineering (Kale and Kulkarni, 2018; Kale and Kulkarni, 2021), non-linear test problems (Zhang et al., 2016), Single Knapsack Problem (Kulkarni and Shabir, 2016) and Multidimensional Knapsack Problem (Poonawala et al., 2024). The FAGA algorithms are implemented in Python 3, and simulations are conducted on a Windows platform. Additionally, each individual problems are run 30 times to ensure robustness. The solutions obtained from the proposed algorithm, along with comparisons to existing algorithms, are discussed in Sect 7.

---

## 5. Static Penalty Function (SPF)

In a constrained optimization problem (Li et al., 2011; Yu et al., 2010), the objective is to minimize or maximize a function:

$$f(X) = (x_1, x_2, \dots, x_n) \tag{5.1}$$

Subject to a set of constraints, which may include both inequality and equality constraints:

$$g_i(X) \leq 0, \quad i = 1,2,\ldots,m \tag{5.2}$$
$$h_j(X) = 0, \quad j = 1,2,\ldots,p \tag{5.3}$$
where:
$g_i(X)$ are the inequality constraints,
$h_j(X)$ are the equality constraints.

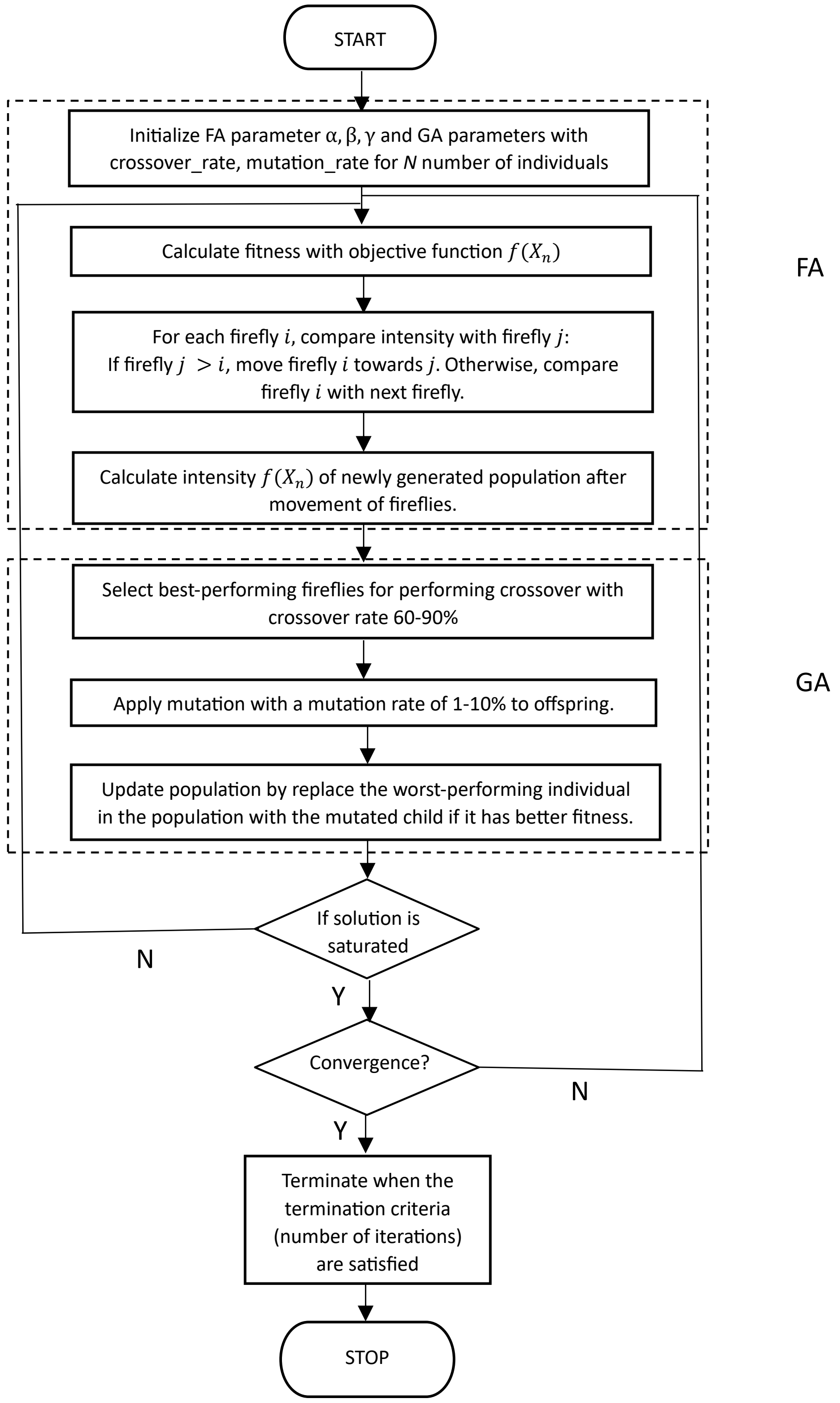

**Fig. 3.** Flow Chart FAGA

To handle constraint violations, a Static Penalty Function (SPF) approach (Kale and Kulkarni, 2021) is commonly used. This approach adds a penalty term, $PF$, to the objective function to discourage violations of both inequality and equality constraints. The penalty function is defined as:

$$PF = \theta \times \left(\sum_{i=1}^{m} g_i(X) + \sum_{j=1}^{p} h_j(X)\right) \quad (5.4)$$

where:

$\theta$ is a constant penalty parameter that controls the intensity of the penalty,

$(\sum_{i=1}^{m} g_i(X) + \sum_{j=1}^{p} h_j(X))$ is summation of violated constraints.

In this formulation, if all constraints are satisfied, both terms in the summation will be zero, resulting in no penalty. However, if there are violations, the penalty $PF$ increases in proportion to the degree of violation.

The penalized objective function is then given by:

$$\phi(X) = f(X) + PF \quad (5.5)$$

where $\phi(X)$ represents the objective function adjusted with the penalty term. This penalized objective function guides the optimization process by imposing a high cost for constraint violations, thereby encouraging feasible solutions that satisfy both the inequality and equality constraints.

## 6. 0-1 Knapsack Problem

The 0-1 Knapsack Problem (KP) (Martello and Toth, 1990; Poonawala et al., 2024; Kulkarni and Shabir, 2016) is a well-known combinatorial optimization problem where the objective is to select items with the maximum possible value without exceeding a given weight limit. Each item either be chosen (1) or drop (0), which gives the problem its "0-1" nature. The KP is categories into the Single Knapsack Problem (SKP), which involves a single knapsack, and the Multidimensional Knapsack Problem (MKP), where multiple constraints are present, such as different weight or size limits. The 0-1 KP has both theoretical and practical significance, with applications in areas like resource allocation, finance, and logistics. In this paper, different variations of the 0-1 knapsack problem are solved by the proposed algorithm. The performance of the FAGA hybrid algorithm is evaluated by examining solution accuracy, convergence speed, and robustness. 0-1 Knapsack problem formulation and the methodology for solving using the FAGA hybrid are illustrated in the following sections.

### 6.1 Problem Formulation

The 0-1 Knapsack Problem involves selecting a subset of items, each having a specific weight $w_i$ and profit $v_i$, with the goal of maximizing the total profit while ensuring that the total weight does not exceed the knapsack's capacity. The task is to identify which items to include so that the total weight stays within the allowed limit, and the combined profit is as high as possible. Various versions of the 0-1 Knapsack Problem exist, where each item is either included in the knapsack ($x_i$ = 1) or excluded ($x_i$= 0). The mathematical formulation for this problem is:

$$Maximize \;\; Z = \sum_{i=1}^{n} v_i x_i \quad (6.1.1)$$

$$subject\ to \;\; \sum_{i=1}^{n} w_i x_i \;\leq W \;, x_i \in \{0, 1\}, \forall_i = 1, \ldots, n \quad (6.1.2)$$

where,

$Z$ represents the total profit to be maximized.

$v_i$ is the profit of item $i$.

$w_i$ is the weight of item $i$.

$x_i$ is a binary decision variable

$W$ is the maximum capacity of the knapsack.

The objective is to maximize the total profit $Z$, subject to the constraint that the total weight of the selected items does not exceed $W$. Each item is either included or excluded from the solution, reflecting the binary nature of the problem.
The Multidimensional Knapsack Problem (MKP) extends this formulation by involving multiple knapsacks, each with its own capacity, where the goal is to distribute the items across several knapsacks while maximizing the overall profit and adhering to the capacity constraints of each knapsack.

## 6.2 Methodology to solve 0-1 Knapsack Problem using FAGA

The hybrid approach combining the FA and GA effectively optimizes item selection for maximum profit while ensuring that weight constraints are not violated. In this method, each firefly symbolizes a potential solution which is represented as a binary vector that indicates whether an item is selected. Fireflies are attracted to brighter (maximum) values, with their movement directed toward these solutions during iterations. The brightness is based on the fitness of the solution, which, in this case, is the total profit relative to the knapsack's capacity. The FA emphasizes local search by refining promising solutions while still exploring new areas randomly to enhance the selection process. On the other hand, the GA evolves a population of potential solutions through selection, crossover, and mutation. Selection favors the fittest individuals for reproduction, crossover combines the genetic material of parent solutions to create new offspring, and mutation introduces diversity by randomly altering item selections, which prevents stagnation in local optima. By integrating these methodologies, the hybrid FAGA approach harnesses FA's strength in local optimization and GA's ability for broader exploration, allowing for an efficient search for near-optimal solutions to the 0-1 Knapsack Problem while balancing exploration and exploitation. Detail explanation is shown in Fig 4 and mathematically is expressed as follows:
***Step 1:*** Each individual in the population represents a solution to the knapsack problem, where an individual is a binary vector $x$ = $[x_1, x_2, \ldots, x_n]$ with $x_i = 1$ if the item is selected and $x_i = 0$ otherwise. The size of the population, $N$ , is predefined, and each solution vector is initialized randomly as:

$$x_i = round\big(lb_i + rand(N) \times (ub_i - lb_i)\big) \tag{6.2.1}$$

Where:
$lb_i$ and $ub_i$ are the lower and upper bounds, respectively (typically $lb = 0$ and $ub = 1$ for the 0-1 Knapsack problem).
$rand(N)$ generates random numbers between 0 and 1, ensuring diversity in the initial population.
***Step 2:*** The sorting of items is based on their profit-to-weight ratio to prioritize more valuable items. After sorting, each item's value is normalized: if the item's value after sorting is less than 0.5, it is set to 0 (unselected); if greater than or equal to 0.5, it is set to 1 (selected). The normalization for a fitness value $f(x_i)$ of an individual is done by scaling the fitness to arrange between 0 and 1. This is mathematically expressed as:

$$f_{norm}(x_i) = \frac{f(x_i) - f_{min}}{f_{max} - f_{min}} \tag{6.2.2}$$

Where:

$$x_i = \begin{cases} 0 & if\ x_i < 0.5 \\ 1 & if\ x_i \geq 0.5 \end{cases} \tag{6.2.3}$$

Here, $f_{min}$ and $f_{max}$ are the minimum and maximum fitness values in the population, and $x_i$ is the normalized selection for item $i$.

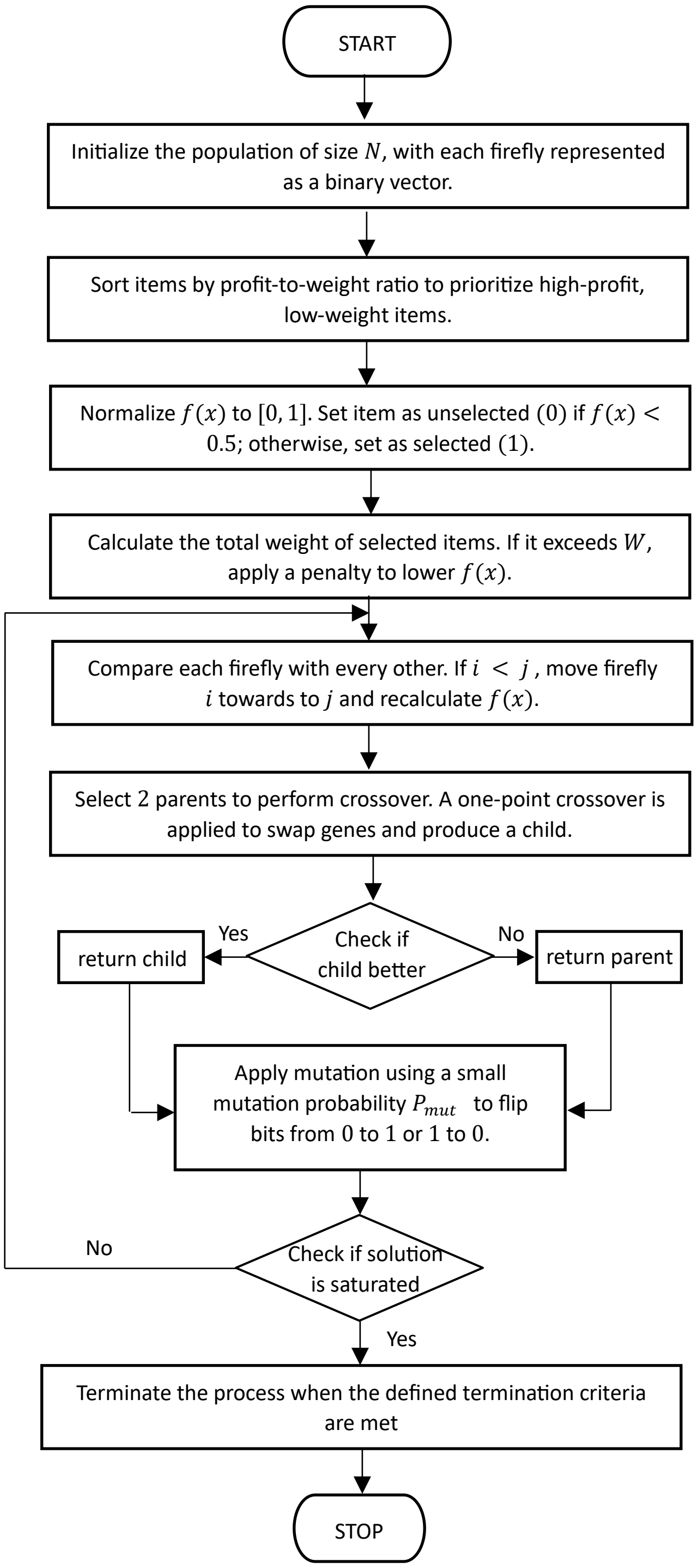

**Fig 4.** Flowchart of solving 0-1 KP using FAGA

***Step 3:*** The fitness function for each individual is calculated based on the profit and weight of selected items. If the total weight exceeds the knapsack capacity $W$, a penalty is applied. The fitness function $f(x)$ is defined as:

$$f(x) = \sum p_i x_i \times \left(\theta \times max(0, \sum w_i x_i - W)\right) \qquad (6.2.4)$$

Where:

$p_i$ is the profit of item $i$,

$w_i$ is the weight of item $i$,

$W$ is the knapsack capacity,

$\theta$ is the static penalty parameter.

***Step 4:*** In the Firefly Algorithm (FA), each individual (firefly) is attracted to brighter (maximum) fireflies. The movement of a firefly $i$ towards a more attractive firefly is governed by the equation (2.2).

***Step 5:*** After the firefly movement, two individuals (parents) are selected randomly for crossover to produce child. A one-point crossover is performed where a random crossover point is chosen, and the genes are swapped between parents to create the child:

$$Child = Parent1[:crossover_{point}] + Parent2[crossover_{point}:] \quad (6.2.5)$$

If the child solution has better fitness than both parents, the child is kept; otherwise, the parents are retained.

***Step 6:*** Mutation is applied to the solution returned from the crossover step. A small mutation probability $P_{mut}$ is used to flip some of the bits (change 0 to 1 or 1 to 0) in the binary solution. This helps introduce diversity and prevents premature convergence. And it is expressed as:

$$x_i' = \begin{cases} 1 - x_i & if\ random() < P_{mut} \\ x_i & otherwise \end{cases} \quad (6.2.6)$$

Where $x_i'$ is the mutated value.

***Step 7:*** The algorithm iterates over the steps of FA movement, GA crossover, and mutation until a predefined stopping criterion is met. This criterion is a fixed number of iterations or converges to an optimal solution.

---

## 7. Results and discussion

The FAGA algorithm is applied to solve four nonlinear benchmark test problems (Zhang et al., 2016; Qi et al., 2017), including both convex and non-convex functions (Diamond et al., 2018) with continuous variables. In addition, the algorithm addresses five mixed-variable design engineering problems (Cheng et al., 2014; Kale and Kulkarni, 2018; Kale and Kulkarni, 2021) and several distinct binary (0-1) knapsack problems (Martello and Toth, 1990; Poonawala et al., 2024; Kulkarni and Shabir, 2016). To handle the constraints in these challenges, a static penalty function (Kale and Kulkarni, 2021) is used. This function applies penalties to solutions that violate constraints, with a fixed penalty proportional to the severity of the violations. This guides the search towards feasible solutions within the search space. In this study, the implementation of the penalty function and the optimization of the FAGA hybrid algorithm are executed using Python libraries for efficient array processing. Additionally, standalone FA and GA are applied to these problems to evaluate the performance of the hybrid FAGA algorithm compared to both algorithms individually.

### 7.1. Benchmark Functions

Benchmark problems (Zhang et al., 2016; Qi et al., 2017) are fundamental methods for evaluating and comparing the performance of optimization algorithms. In this study, the proposed algorithm aims to solve four widely recognized benchmark functions: the Sphere, Ackley, Rosenbrock, and Rastrigin functions (Qi et al., 2017), as shown in Table 1. These functions represent a variety of optimization challenges, each characterized by its complexity, modality, and search space landscape. The Sphere function is simple and unimodal, used to evaluate an algorithm's ability to perform basic optimization. The Ackley function, Rosenbrock function and Rastrigin function are multimodal and difficult due to their many local minima, making it challenging for algorithms to avoid getting trapped. The proposed FAGA algorithm is compared against several state-of-the-art optimization algorithms, such as Differential Evolution (DE) (Zhang et al., 2016), Particle Swarm Optimization (PSO) (Zhang et al., 2016), Hybrid Firefly Algorithm (HFA) (Zhang et al., 2016), Genetic

Algorithm (GA) (Qi et al., 2017), and the standard Firefly Algorithm (FA) (Zhang et al., 2016). The goal of these comparisons is to evaluate how well FAGA performs in terms of convergence speed, solution accuracy, and overall robustness. By testing FAGA against these benchmark functions, algorithm determine its effectiveness in addressing a broad range of optimization problems (Li et al., 2011; Yu et al., 2010), from simple to highly complex landscapes which mimic real-world optimization challenges.

**Table 1.** Test Function Formulation and Parameters

| Function | Range | Type | Formulation | Global minimum |
|---|---|---|---|---|
| Sphere | [-5.12, 5.12] | U | $f_1 = \sum_{i=1}^{n} x_i^2$ | 0 |
| Ackley | [-15,30] | M | $f_2 = 20 + e - 20e^{-0.2}\sqrt{\frac{1}{n}\sum_{i=1}^{n}\cos(2\pi x_i)}$ | 0 |
| Rosenbrock's | [-5,10] | M | $f_3 = \sum_{i=1}^{n-1}(100(x_{i+1} - x_1^2)^2 + (x_i - 1)^2)$ | 0 |
| Rastrigin | [-5.12, 5.12] | M | $f_4 = \sum_{i=1}^{n}(x_i^2 - 10\cos(2\pi x_i) + 10)$ | 0 |

(U = Unimodal; M = Multimodal)

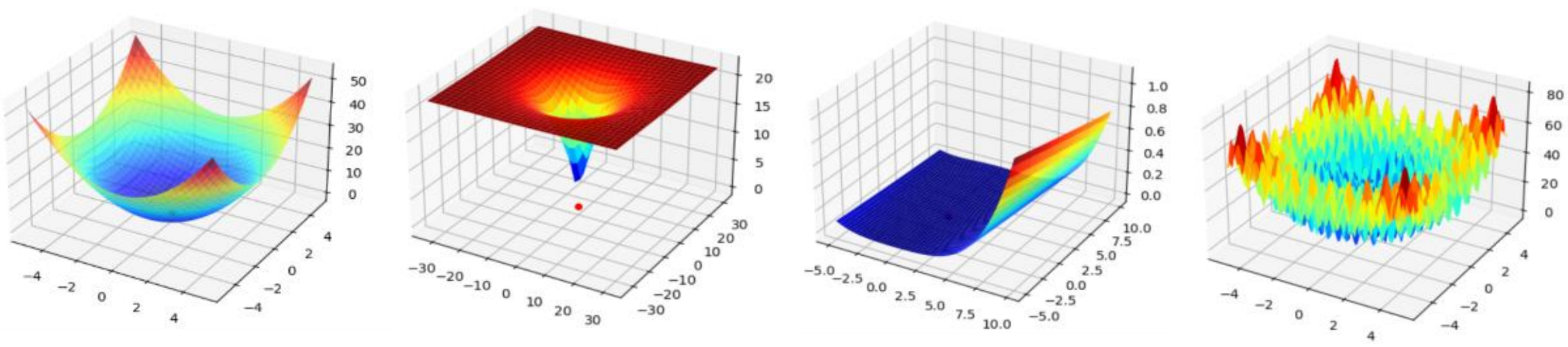

**Sphere** **Ackley** **Rosenbrock's** **Rastrigin**

**Fig 5.** Perspective view of test functions

Table 2, presents the statistical performance of FAGA evaluated over 30 trials for each benchmark function, with dimension 30. The trial approach ensures robust results and accounts for the stochastic nature of the algorithm. FAGA consistently produces competitive results, achieving near-optimal solutions with low variance. For instance, in the Sphere and Ackley functions, which have well-defined global minima, FAGA effectively minimizes the objective function to values close to zero. Furthermore, the algorithm demonstrates resilience in addressing challenging functions such as Rastrigin and Rosenbrock, effectively avoiding local minima and steadily converging toward the global minimum. Statistical metrics, including best, worst, mean, standard deviation, function evaluations, and computational time, highlight the algorithm's efficiency and adaptability across diverse optimization problems.

In Table 3, FAGA's comparative performance with other algorithms is thoroughly analyzed across all four benchmark functions. When compared with DE (Zhang et al., 2016), HFA (Zhang et al., 2016), PSO (Zhang et al., 2016), GA (Qi et al., 2017), and FA (Zhang et al., 2016), FAGA consistently outperforms most algorithms. It particularly excels over PSO and FA by achieving mean values closer to the optimal solution and best values in the Sphere function, showcasing its strong ability to efficiently optimize smooth, unimodal landscapes (Garden et al., 2014). In the Ackley function, FAGA also demonstrates superior performance, especially in terms of convergence speed and avoiding local minima, outperforming both FA and DE.

**Table 2.** Performance analysis of test functions using FAGA

| Statistics | Sphere | Ackley | Rosenbrock' | Rastrigin |
|---|---|---|---|---|
| **Best** | 4.06E-117 | 1.27E-16 | 2.04E-15 | 8.45E-01 |
| **Worst** | 8.79E-99 | 4.41E-16 | 5.67E-08 | 1.05E+00 |
| **Mean** | 6.73E-100 | 2.90E-16 | 4.39E-09 | 9.42E-01 |
| **Std. Dev** | 1.873E-99 | 1.01E-16 | 1.06E-08 | 0.0563 |
| **Avg. Fun. Eval.** | 499103 | 71626 | 508058 | 397387 |
| **Avg. CPU time (s)** | 1.07E+02 | 1.16E+02 | 1.14E+02 | 1.73E+02 |
| **Total CPU time (s)** | 3223.43 | 3485.86 | 158.64 | 5182.09 |

**Table 3.** Comparison of FAGA with various algorithms for test functions

| Function | Statistics | HFA (Zhang et al., 2016) | DE (Zhang et al., 2016) | PSO (Zhang et al., 2016) | FA (Zhang et al., 2016) | GA (Qi et al., 2017) | FAGA |
|---|---|---|---|---|---|---|---|
| $f_1$ | **Min** | 1.07E-193 | 1.395e-09 | 6.33E-13 | 1.02E-87 | NA | 4.06E-117 |
| | **Max** | 7.84E-170 | 6.127e-08 | 6.86E-10 | 1.95E-87 | NA | 8.79E-99 |
| | **Mean** | **2.64E-171** | 1.416e-08 | 9.43E-11 | 1.57E-87 | 1.34e+00 | 6.73E-100 |
| | **Std** | 0 | 1.295e-08 | 1.48E-10 | 1.89E-88 | 4.09e-001 | 1.873E-99 |
| $f_2$ | **Min** | 4.44E-15 | 4.363e-05 | 4.75E-07 | 7.99E-15 | NA | 1.27E-16 |
| | **Max** | 6.13E-05 | 0.003 | 6.15E-05 | 1.51E-14 | NA | 4.41E-16 |
| | **Mean** | 1.31E-05 | 3.127e-04 | 7.10E-06 | 1.25E-14 | 1.98e+001 | **2.90E-16** |
| | **Std** | 2.33E-05 | 5.487e-04 | 1.47E-05 | 3.36E-15 | 3.84e-001 | 1.01028E-16 |
| $f_3$ | **Min** | 2.47E-29 | 14.912 | 1.876 | 26.346 | NA | 2.04E-15 |
| | **Max** | 0.530 | 25.267 | 114.49 | 89.131 | NA | 5.67E-08 |
| | **Mean** | 0.077 | 21.999 | 49.686 | 29.053 | 1.39e+003 | **4.39E-09** |
| | **Std** | 0.161 | 2.103 | 34.029 | 11.348 | 8.14e+002 | 1.06E-08 |
| $f_4$ | **Min** | 1.08E-08 | 143.889 | 17.909 | 3.979 | NA | 8.45E-01 |
| | **Max** | 4.36E-08 | 196.363 | 44.773 | 15.919 | NA | 1.05E+00 |
| | **Mean** | **3.39E-08** | 175.911 | 30.15 | 9.386 | 1.57e+002 | 9.42E-01 |
| | **Std** | 7.29E-09 | 12.243 | 7.108 | 3.044 | 5.02e+001 | 0.056334 |

However, in more complex functions like Rosenbrock and Rastrigin, the performance gap between FAGA and the other algorithms narrows. While FAGA still manages to find reasonably good solutions, HFA (Zhang et al., 2016) at times delivers better results in terms of effectively exploring the solution space. Overall, FAGA maintains a competitive edge by offering balanced performance. The convergence of FAGA, shown in Fig. 6, is analyzed by observing the best fitness values over a series of iterations for each benchmark function. The convergence plots demonstrate that FAGA achieves rapid convergence in simpler functions like Sphere and Ackley, reaching near-optimal solutions within a few hundred iterations. For more complex functions like Rosenbrock and Rastrigin, FAGA exhibits steady convergence, though at a slower rate due to the complexity of these functions. This gradual approach ensures that FAGA avoids premature convergence while continuously improving the solution quality over time.

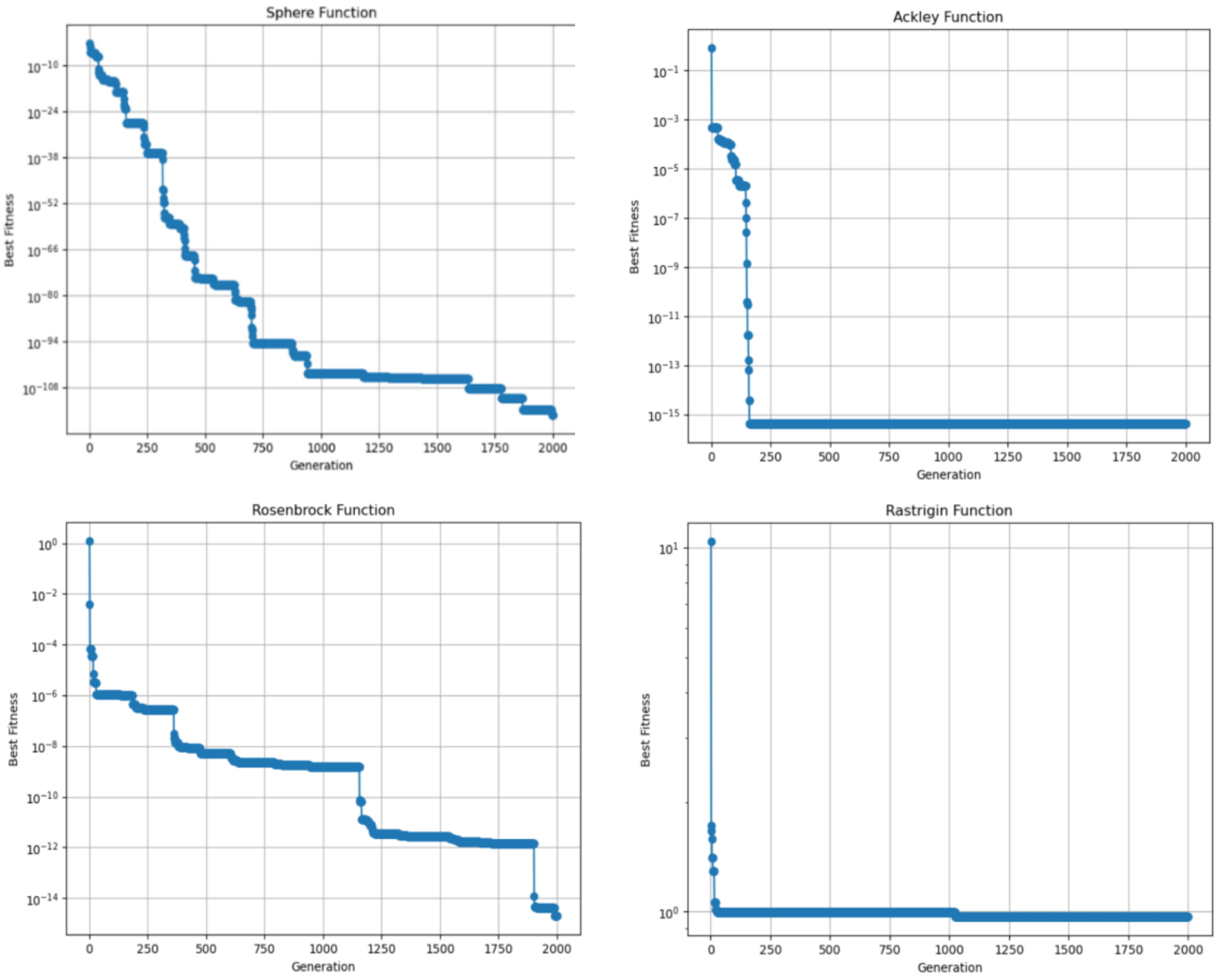

**Fig 6.** Convergence results of test functions using FAGA

## 7.2 Design engineering problems

The effectiveness of the proposed FAGA hybrid algorithm is demonstrated through five complex engineering design challenges (Kale and Kulkarni, 2021): the cantilever beam problem (aimed at minimizing weight), the I-beam vertical deflection problem (focused on reducing deflection), the gear train problem (which seeks to minimize gear ratios), the pressure vessel problem (minimization of cost), and the coil compression problem (designed to minimize volume). Each of these problems involves a combination of continuous and discrete design variables. For statistical analysis and performance assessment, the FAGA hybrid, as well as the individual FA and GA algorithms, are executed 30 times for each design problem.

**Test example-1: Helical Compression Spring Design**

The mixed-variable problem of designing a helical compression spring, made of alloyed steel, is illustrated in Figure 7. It involves both discrete and continuous variables, with the objective of minimizing the volume (V) of the spring. The formulation of the problem is presented below:

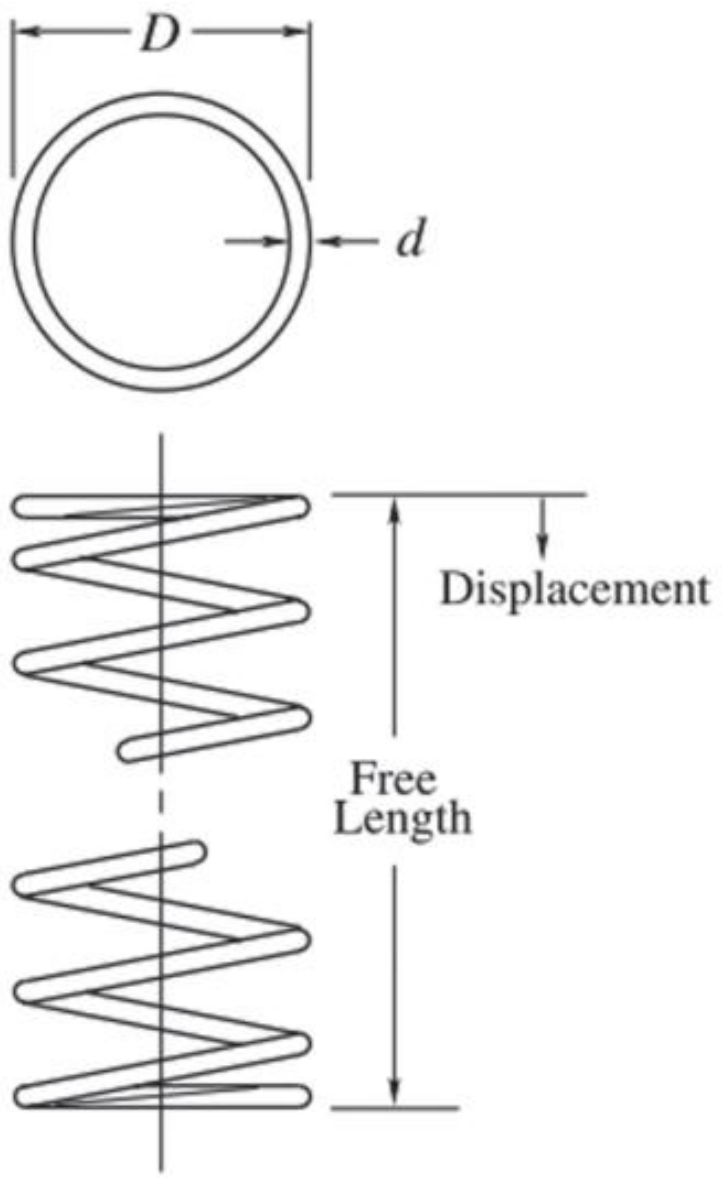

**Fig** 7. Helical Compression Spring Design Problem

$$Minimise\ f = V = \frac{\pi^2 D d^2 (N+2)}{4} \tag{7.2.1}$$

$$subject\ to\ g_1 = \frac{8KP_{max}D}{\pi d^3} - S \leq 0 \tag{7.2.2}$$

$$g_2 = \left(\frac{P_{max}}{k} - 1.05(N+2)d\right) \tag{7.2.3}$$

$$g_3 = d_{min} - d \leq 0 \tag{7.2.4}$$

$$g_4 = (d + D) - D_{max} \leq 0 \tag{7.2.5}$$

$$g_5 = 3 - \frac{D}{d}\ 0 \tag{7.2.6}$$

$$g_6 = \delta_p - \delta_{pm} \leq 0 \tag{7.2.7}$$

$$g_7 = \left(\frac{P_{max}}{k} - 1.05(N+2)d - L_f\right) - L_{free}\ 0 \tag{7.2.8}$$

$$g_8 = \delta_w - \left(\frac{P_{max} - P}{k}\right) \leq 0 \tag{7.2.9}$$

**Table 4.** Specific design sizes for the wire diameter *d*

| | | | | |
|---|---|---|---|---|
| 0.0090 | 0.0162 | 0.0350 | 0.1050 | 0.2250 |
| 0.0095 | 0.0173 | 0.0410 | 0.1200 | 0.2440 |
| 0.0104 | 0.0180 | 0.0470 | 0.1350 | 0.2830 |
| 0.0118 | 0.0200 | 0.0540 | 0.1620 | 0.3070 |
| 0.0128 | 0.0230 | 0.0720 | 0.1770 | 0.3310 |
| 0.0132 | 0.0280 | 0.0800 | 0.1920 | 0.3620 |
| 0.0150 | 0.0320 | 0.0920 | 0.2070 | 0.3940 |

The specific design size for the wire diameter $d$, as presented in Table 4, is taken **0.2830** for the proposed algorithms. Table 5 provides the statistical outcomes for the helical compression spring problem for FA, GA, and FAGA. FAGA achieves the best function value, matching FA and surpassing GA. The mean and worst function values further validate FAGA consistency. Notably, FAGA significantly reduces the average computational time, compared to FA and GA.

**Table 5.** Statistical results compression helical spring problem using FAGA

| Results | FA | GA | FAGA |
|---|---|---|---|
| **Best** | 2.66 | 2.82 | 2.66 |
| **Mean** | 2.66 | 2.98 | 2.66 |
| **Worst** | 2.67 | 3.2 | 2.67 |
| **Std. Dev** | 0.194e-2 | 0.132 | 0.187e-2 |
| **Avg. CPU time (sec)** | 10.47 | 2.78 | 0.95 |
| **Avg. Fun. Eval.** | 14728 | 22707 | 7787 |

In Table 6, the performance of FAGA is compared with other algorithms, such as Nonlinear B&B (Sandgren, 1990), AHGA (Yun, 2005), PC (Kulkarni et al., 2016), MRSLS (Kale and Kulkarni, 2021) and CBO (Kale and Kulkarni, 2021). FAGA delivers an optimal spring volume of **2.6586**, matching FA and PC, while outperforming GA, Nonlinear B&B, MRSLS, and CBO. FAGA requires significantly fewer function evaluations than GA, FA, and PC. These results affirm FAGA's capability to deliver accurate solutions with reduced computational cost compared to some conventional algorithms.

**Table 6.** Performance of various algorithms for solving helical spring design problem

| Techniques | Nonlinear B&B (Sandgren, 1990) | AHGA (Yun, 2005) | PC (Kulkarni et al., 2016) | MRSLS (Kale and Kulkarni, 2021) | CBO (Kale and Kulkarni, 2021) | FA | GA | FAGA |
|---|---|---|---|---|---|---|---|---|
| $d$ | 0.2830 | 0.2830 | 0.2830 | 0.283 | 2 | 0.2830 | 0.2830 | 0.2830 |
| $D$ | 1.180701 | 1.1096 | 1.2231 | 1.1808 | 0.3310 | 1.2231 | 1.1810 | 1.2231 |
| $N$ | 10 | 9 | 9 | 10 | 4 | 9 | 10 | 9 |
| **Spring volume $f(x)$** | 2.7995 | 2.0283 | 2.6586 | 2.8002 | 3.2439 | 2.6586 | 2.82 | 2.6586 |
| **Function Evaluations** | NA | NA | 498,567 | 2044 | 108 | 12060 | 22456 | 8460 |

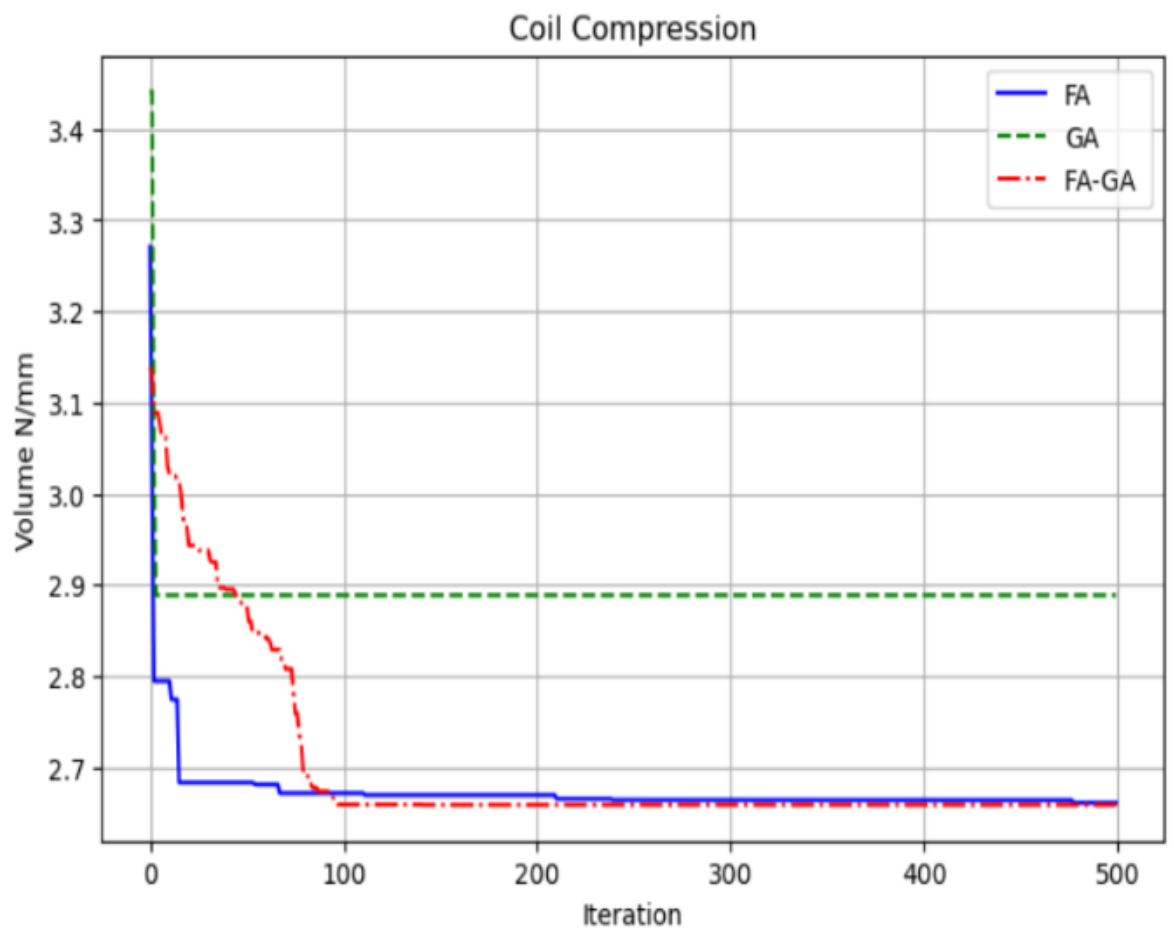

**Fig. 8.** Comparison of the Convergence Curves of FA, GA and FAGA for Solving the Helical Spring Design Problem

Figure 8 illustrates the convergence curves of FA, GA, and FAGA for the helical spring design problem. The graph clearly highlights FAGA's efficient convergence behavior, as it rapidly reaches the optimal solution. Compared to FA, which converges steadily, and GA, which plateaus early with suboptimal results, FAGA demonstrates a balanced approach between exploration and exploitation.

**Test example-2: Pressure Vessel Design Problem**

The optimal design problem for a pressure vessel illustrated in Figure 9, involves both discrete and continuous variables. The discrete variables are the thickness of the spherical head $(x1)$ and the shell thickness $(x2)$, while the continuous variables are the shell's radius $(x3)$ and its length $(x4)$. The formulation of the problem is presented below:

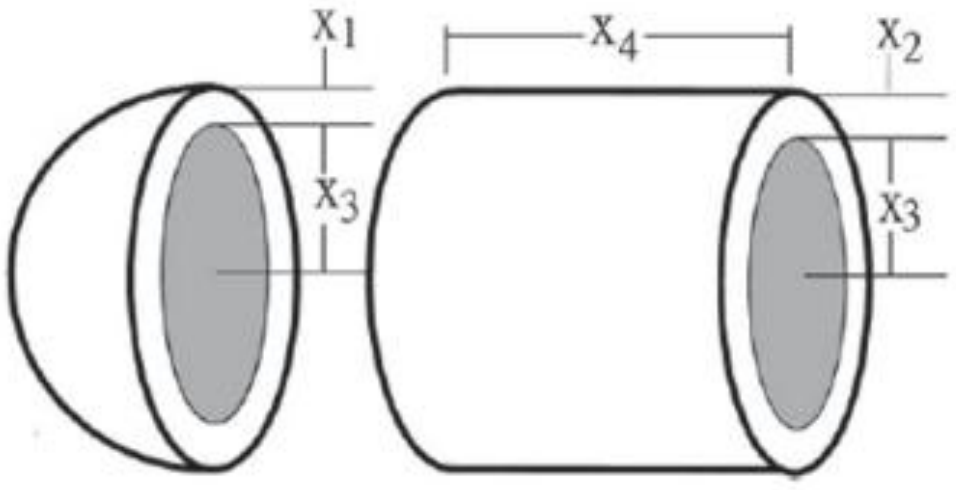

**Fig 9.** Tube and Pressure Vessel

$$Minimise\ f(x) = 0.6224x_1x_3x_4 + 1.7781x_2x_3^2 + 3.1661x_1^2x_4 + 19.84x_1^2x_3 \quad (7.2.10)$$

$$subject\ to\ g(x_1) = -x_1 + 0.0193x_3 \leq 0 \quad (7.2.11)$$

$$g(x_2) = -x_2 + 0.00954x_3 \leq 0 \quad (7.2.12)$$

$$g(x_3) = -\pi x_3^2 x_4 \frac{4}{3}\pi x_3^3 + 750 \times 1782 \leq 0 \quad (7.2.13)$$

$$g(x_4) = -240 + x_4 \leq 0 \quad (7.2.14)$$

$$1 \leq x_1 \leq 1.375 \quad (7.2.15)$$
$$0.625 \leq x_2 \leq 1 \quad (7.2.16)$$
$$48 \leq x_3 \leq 52 \quad (7.2.17)$$
$$90 \leq x_4 \leq 112 \quad (7.2.18)$$

Table 7, presents a statistical comparison of FA, GA and FAGA for the pressure vessel design problem. FAGA achieves the best solution with a cost of **6059.71**, outperforming both FA **6090.92** and GA **6117.8**. Additionally, FAGA demonstrates significantly lower standard deviation compared to GA, indicating more consistent results. However, the hybrid approach shows slightly higher average CPU time compared to GA which remains efficient relative to FA. FAGA also reduces the average number of function evaluations compared to both FA and GA.

**Table 7:** Statistical results pressure vessel design problem

| **Results** | **FA** | **GA** | **FAGA** |
|---|---|---|---|
| **Best** | 6090.92 | 6117.8 | 6059.71 |
| **Mean** | 2.66 | 6268.7 | 6097.75 |
| **Worst** | 2.67 | 6446.26 | 6195.59 |
| **Std. Dev** | 0.194e-2 | 121.50 | 37.6342 |
| **Avg. CPU time (sec)** | 55.009 | 14.206 | 77.991 |
| **Avg. Fun. Eval.** | 248685 | 303468 | 183079 |

In Table 8, a comprehensive performance comparison of multiple algorithms for the pressure vessel design problem, including CPSO (He and Wang, 2007), MRSLS (Kale and Kulkarni, 2021), LCA (Kashan, 2011), OIO (Kashan,2015), CI-SPF (Kale and Kulkarni, 2018), GA and FA. FAGA achieves the best cost value of **6059.72**, matching the optimal results of LCA, OIO, and CI-SPF. Additionally,

FAGA outperforms MRSLS, CPSO, GA, and FA, demonstrating its superior efficiency in solving the problem. Across the design variables $x1$, $x2$, $x3$, and $x4$, FAGA closely aligns with CPSO and CI-SPF. Furthermore, FAGA demonstrates efficiency in function evaluations, which is significantly fewer than GA and FA, highlighting its computational advantage.

**Table 8.** Performance of various algorithms for solving pressure vessel design problem

| Techniques | CPSO (He and Wang, 2007) | LCA (Kashan, 2011) | OIO (Kashan, 2015) | CI–SPF (Kale and Kulkarni, 2018) | MRSLS (Kale and Kulkarni, 2021) | GA | FA | FAGA |
|---|---|---|---|---|---|---|---|---|
| $x1$ | 0.8125 | NA | NA | 0.8125 | 0.8125 | 0.8125 | 0.8125 | 0.8125 |
| $x2$ | 0.4375 | NA | NA | 0.4375 | 0.4375 | 0.4375 | 0.4375 | 0.4375 |
| $x3$ | 42.09126 | NA | NA | 42.0984 | 41.9645 | 43.2146 | 40.9929 | 42.0923 |
| $x4$ | 176.7465 | NA | NA | 176.6366 | 178.3043 | 163.2884 | 190.8427 | 176.8701 |
| Cost $f(x)$ | 6061.08 | 6059.85 | 6059.71 | 6059.72 | 6076.12 | 6117.8 | 6090.92 | 6059.72 |
| Function Evaluations | 200000 | 24000 | 50000 | 124581 | 1200 | 329874 | 287134 | 162460 |

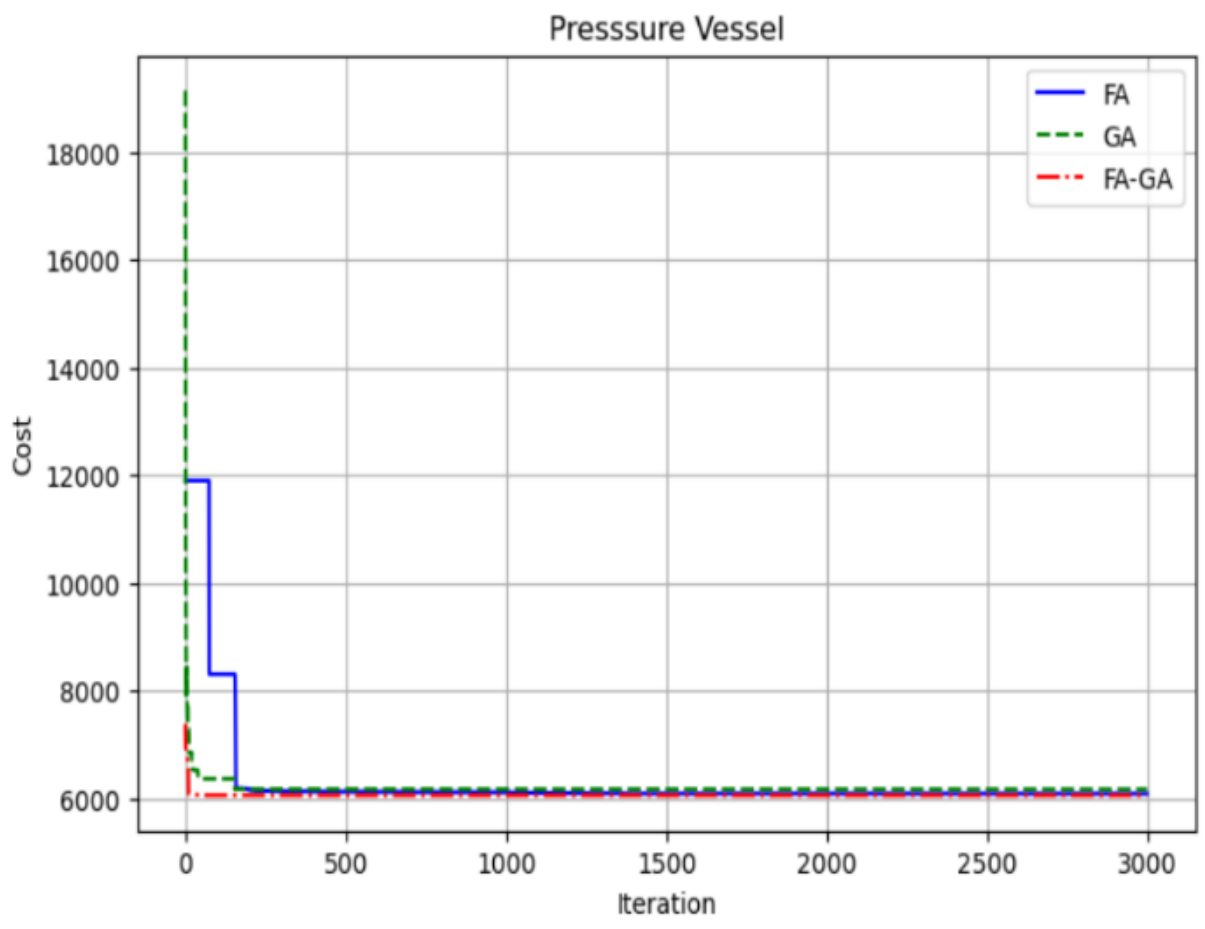

**Fig 10.** Comparison of the Convergence Curves of FA, GA and FAGA for Solving the pressure vessel design problem

Figure 10, illustrates the convergence curves of FA, GA, and FAGA for the pressure vessel design problem. The FAGA algorithm achieves early convergence with minimal cost fluctuation, demonstrating its ability to exploit the search space efficiently. In contrast, the FA curve shows more exploration during the initial iterations, delaying convergence. GA struggles with slower convergence and does not achieve the same performance level as FAGA. Overall, the figure highlights FAGA's ability to balance exploration and exploitation, resulting in faster convergence and better solution quality compared to FA and GA.

## Test example-3 cantilever beam design problem

The cantilever beam problem involves a cantilever beam consisting of five elements, each with a hollow cross-section of fixed diameter, as shown in Figure 11. The beam is rigidly supported, and a vertical force is applied at the free end. The objective of the problem is to minimize the weight of the beam. The design variable is the height (or width) $x_i$ of each beam element. The formulation of the problem is presented below:

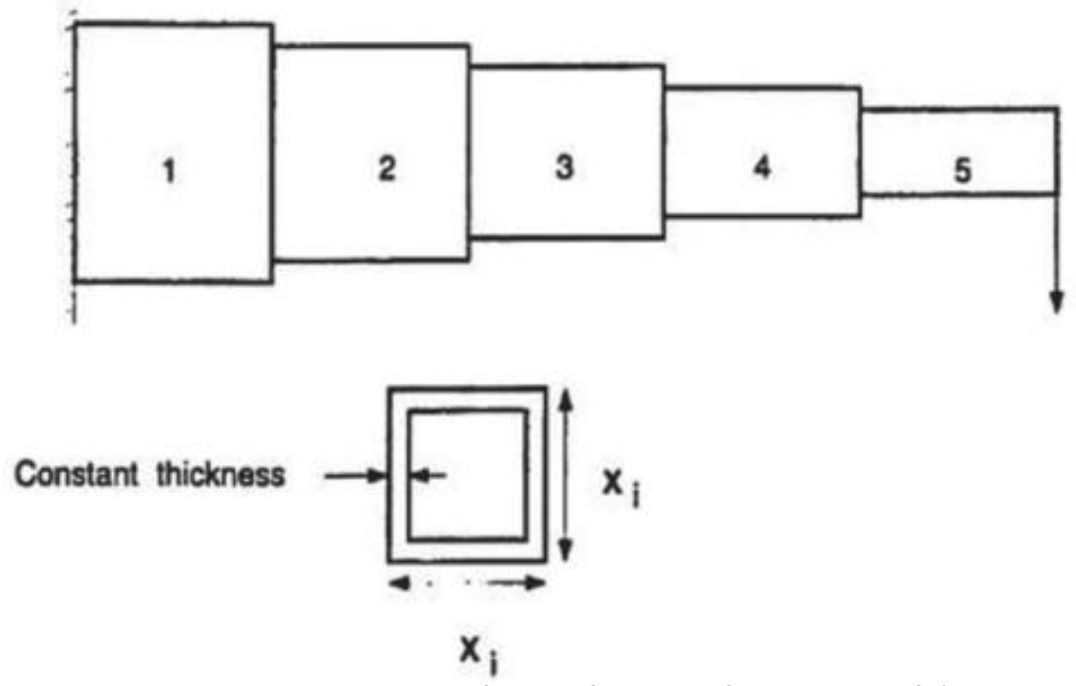

**Fig 11.** cantilever beam design problem

$$Minimise\ f(x) = 0.0624(x_1 + x_2 + x_3 + x_4 + x_5) \tag{7.2.19}$$

$$subject\ to\ g_1 = \frac{61}{x_1^3} + \frac{37}{x_2^3} + \frac{19}{x_3^3} + \frac{7}{x_4^3} + \frac{1}{x_5^3} \leq 1 \tag{7.2.20}$$

$$0.01 \leq x_i \leq 100 \tag{7.2.21}$$

In Table 9, the statistical results for FA, GA, and FAGA in solving the cantilever beam design problem. The best solution achieved by FAGA matches the best result from FA and surpasses GA. Additionally, FAGA demonstrates a mean value closer to the optimal solution compared to GA's. The standard deviation for FAGA is the lowest, indicating higher consistency and reliability in the solutions. FAGA also shows a balanced computational efficiency outperforming GA and FA.

**Table 9.** Statistical results cantilever beam design problem

| **Results** | **FA** | **GA** | **FA – GA** |
|---|---|---|---|
| **Best** | 1.3399 | 1.3422 | 1.3399 |
| **Mean** | 1.3421 | 1.3446 | 1.3408 |
| **Worst** | 1.349 | 1.349 | 1.3443 |
| **Std. Dev** | 0.003 | 0.0026 | 1.36E-3 |
| **Avg. CPU time (sec)** | 10.321 | 3.667 | 5.7447 |
| **Avg. Fun. Eval.** | 3018 | 7410 | 4709 |

In Table 10, the performance of FAGA with various other optimization algorithms, including MRSLS (Kale and Kulkarni, 2021), CBO (Kale and Kulkarni, 2021), CS (Gandomi et al., 2013), SOS (Cheng and Prayoga, 2014), and CI-SAPF-CBO (Kale and Kulkarni, 2021). FAGA achieves a weight value **1.3399** which is identical to the results of FA, CS, CI-SAPF-CBO and SOS while outperforming other techniques such as MRSLS and CBO. Additionally, FAGA achieves this result with fewer function evaluations, which is considerably better than GA and SOS. The performance of FAGA demonstrates its ability to achieve optimal solutions efficiently and consistently while maintaining robustness.

Figure 12, illustrates the convergence curves of FA, GA, and FAGA for the cantilever beam design problem. The FAGA curve shows the fastest convergences well as maintaining stability. GA converges quickly although displays slight fluctuations before stabilizing. FA on the other hand, converges more slowly. This figure highlights the superiority of FAGA, as it combines the rapid convergence of GA with the consistency and robustness of FA, resulting in a highly efficient optimization process.

**Table 10.** Performance of various algorithms for solving cantilever beam design problem

| Techniques | CS (Gandomi et al., 2013) | SOS (Cheng and Prayoga, 2014) | MRSLS (Kale and Kulkarni, 2021) | CBO (Kale and Kulkarni, 2021) | CI-SAPF-CBO (Kale and Kulkarni, 2021) | FA | GA | FAGA |
|---|---|---|---|---|---|---|---|---|
| **x1** | 6.0089 | 6.01878 | 5.9356 | 12.4548 | 6.0064 | 6.0309 | 6.0216 | 6.0271 |
| **x2** | 5.3049 | 5.30344 | 5.2700 | 5.4801 | 5.3134 | 5.3132 | 5.3757 | 5.2938 |
| **x3** | 4.5023 | 4.49587 | 4.5587 | 8.3861 | 4.4983 | 4.4859 | 4.3931 | 4.4971 |
| **x4** | 3.5077 | 3.49896 | 3.5333 | 19.8751 | 3.4952 | 3.4989 | 3.6025 | 3.4981 |
| **x5** | 2.1504 | 2.15564 | 2.1932 | 5.0897 | 2.1602 | 2.1450 | 2.1177 | 2.1580 |
| **Weight f(x)** | 1.3399 | 1.3399 | 1.3410 | 3.2002 | 1.3399 | 1.3399 | 1.3422 | 1.3399 |
| **Function Evaluations** | NA | 15000 | 680 | 2190 | 3025 | 3789 | 9210 | 6000 |

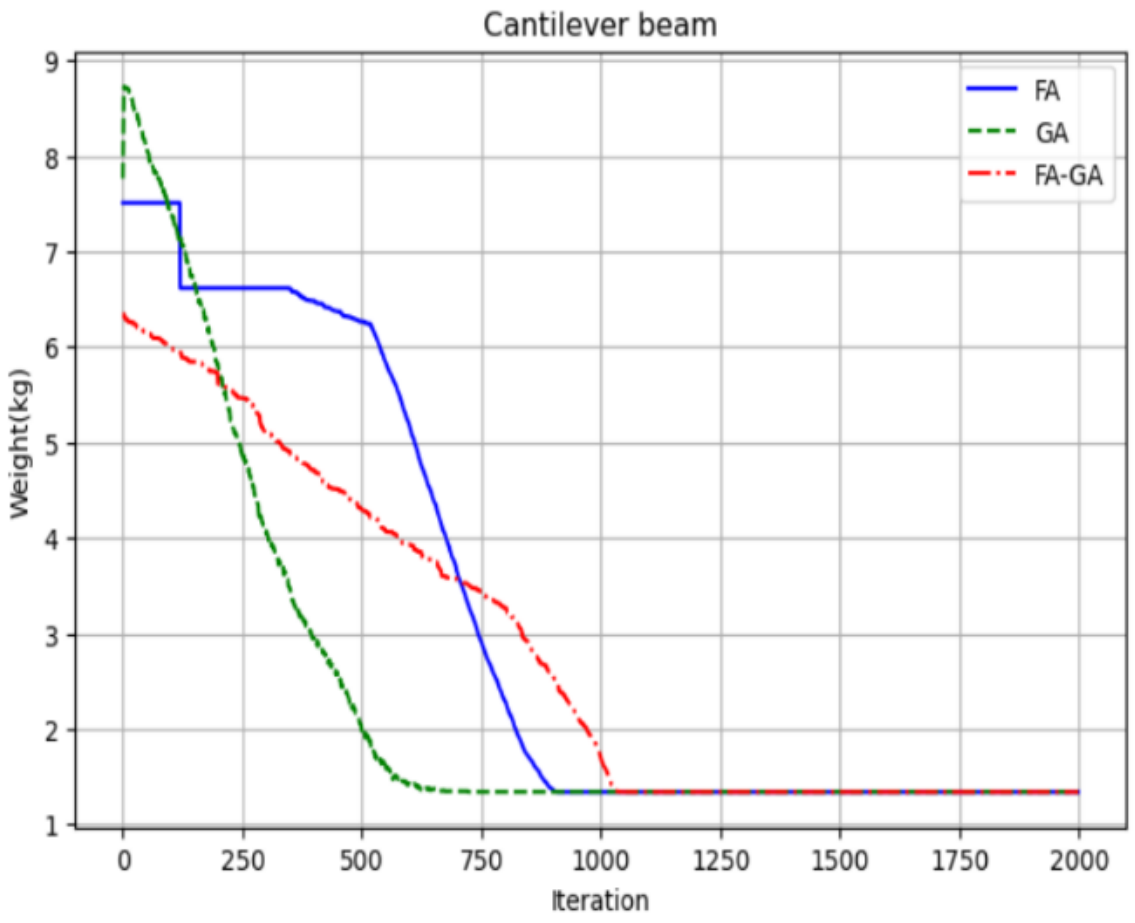

**Fig. 12.** Comparison of the Convergence Curves of FA, GA and FAGA for Solving the cantilever beam design problem

## Test example-4: Gear train design problem

Gear train design problem focuses on optimizing the gear ratio of a compound gear train to efficiently transmit the desired motion or power between two shafts. The gear train, illustrated in Figure 13, includes two pairs of gearwheels: $a$, $b$, $c$ and $d$, where $a$ and $b$ are the driving gears, and $c$ and $d$ are the driven gears. The overall gear ratio is defined as:

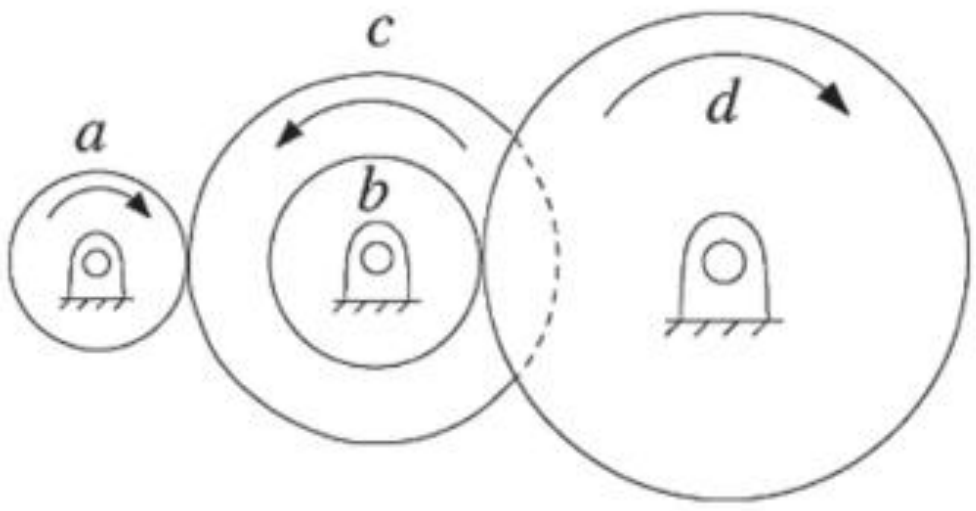

**Fig 13.** Gear Train

$$Gear\ Ratio = \frac{\Pi(\text{Teeth on Driving Gears})}{\Pi(\text{Teeth on Driving Gears})} = \frac{z_a z_b}{z_c z_d} \qquad (7.2.22)$$

where $z_a$, $z_b$, $z_c$ and $z_d$ represent the number of teeth on the respective gears. The goal is to determine the values of $z_a, z_b, z_c$ and $z_d$ that produce a gear ratio close to 1/6.931. The constraint is that the number of teeth on each gear must be within the range {12, 13, ..., 59, 60}. The optimization problem is then formulated below:

$$Minimise\ f(x) = \left(\frac{1}{6.931} - \frac{z_a z_b}{z_c z_d}\right)^2 \tag{7.2.23}$$

$$subject\ to\ g(x) = 12 \leq z_a, z_b, z_c\ , z_d \leq 60 \tag{7.2.24}$$

**Table 11.** Statistical results Gear train design problem

| Results | FA | GA | FAGA |
|---|---|---|---|
| **Best** | 2.7E-12 | 2.3E-11 | 2.7E-12 |
| **Mean** | 4.8E-11 | 8.58E-09 | 8.95E-12 |
| **Worst** | 2.35E-09 | 3.98E-08 | 2.3E-11 |
| **Std. Dev** | 4.94E-10 | 9.96E-09 | 9.55E-12 |
| **Avg. CPU time (sec)** | 2.2334 | 2.3341 | 2.1516 |
| **Avg. Fun. Eval.** | 1041 | 1215 | 632 |

In Table 11, represents the statistical comparison between the FA, GA and the proposed FAGA, for solving the gear train design problem. FAGA achieves the best solution value of $\mathbf{2.7 \times 10^{-12}}$, matching the performance of FA and significantly outperforming GA's $\mathbf{2.3 \times 10^{-11}}$. FAGA demonstrates improved computational efficiency with fewer function evaluation, compared to FA and GA. Furthermore, the Average CPU Time is lowest for FAGA, showing its ability to converge faster while maintaining high precision.

Table 12, highlights the comparative performance of FAGA against other state-of-the-art algorithms, including Lagrange Multiplier (Kannan and Kramer, 1994), PSO (Datta and Figueira, 2011), CBO (Kale and Kulkarni, 2021), and CI-SAPF (Kale and Kulkarni, 2021). FAGA achieves a remarkably low ratio $f(x)$, indicating its superiority over algorithms like Lagrange Multiplier, CBO and GA. Additionally, FAGA requires fewer Function Evaluations, which is significantly better than FA, GA and CI-SAPF. This reduction in evaluations further establishes FAGA's robustness and efficiency when compared to traditional and hybrid optimization algorithms.

**Table 12**. Performance of various algorithms for solving Gear train design problem

| Techniques | Lagrange Multiplier (Kannan and Kramer, 1994) | PSO (Datta and Figueira, 2011) | CBO (Kale and Kulkarni, 2021) | CI-SAPF (Kale and Kulkarni, 2021) | FA | GA | FAGA |
|---|---|---|---|---|---|---|---|
| $z1$ | 13 | 16 | 18 | 16 | 16 | 13 | 16 |
| $z2$ | 15 | 19 | 20 | 19 | 19 | 30 | 19 |
| $z3$ | 33 | 43 | 58 | 43 | 43 | 51 | 43 |
| $z4$ | 41 | 49 | 43 | 49 | 49 | 53 | 49 |
| **Ratio $f(x)$** | 2.1246e-08 | 2.7e-12 | 4.5e-09 | 2.7e-12 | 2.7e-12 | 9.96e-09 | 2.7e-12 |
| **Function Evaluations** | NA | NA | 420 | 1260 | 1120 | 1148 | 873 |

The convergence graph of FA, GA, and FAGA, shown in Fig. 14, represents the results for solving the gear train design problem. It is evident that FAGA exhibits early convergence and is successful in finding the optimal solution. While GA demonstrates a similar convergence trend, it fails to reach the optimal solution. FA performs comparably in terms of convergence and achieving the optimal solution, in terms of both precision and efficiency. The curve for FAGA stabilizes early, as the simplicity of the problem limits diversity among the algorithms.

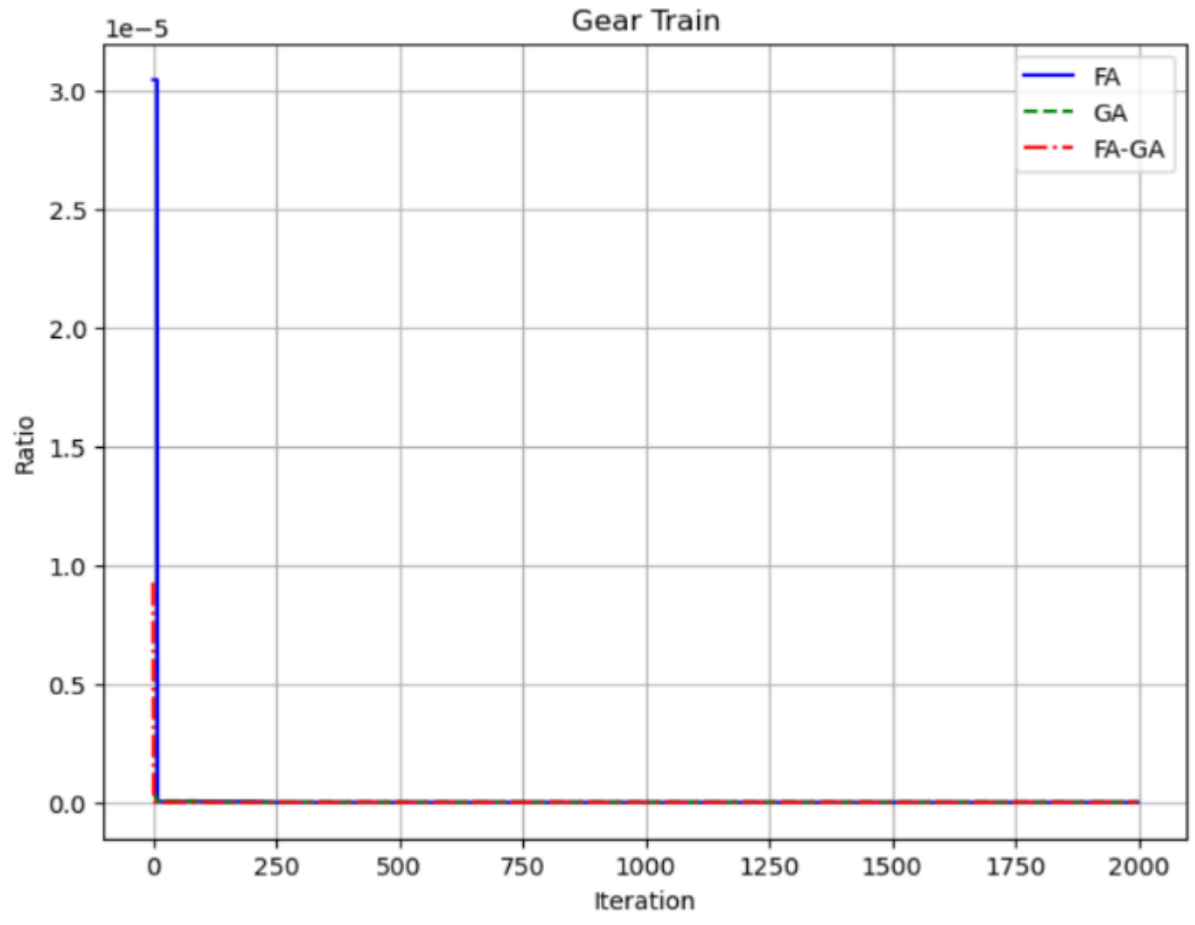

**Fig. 14.** Comparison of the Convergence Curves of FA, GA and FAGA for Solving the Gear train design problem

### Test example-5: I-Beam Vertical Deflection Design Problem

I-beam design problem of minimizing the deflection of an I-beam using four variables. As shown in Figure 15, the objective is to reduce the vertical deflection of the I-beam. This is achieved while satisfying both the cross-sectional area and stress constraints under specific loading conditions. The goal is to minimize the vertical deflection, expressed as $f(x) = \frac{PL^3}{48EI}$, where the beam length $L$ is $5200\,cm$ and the modulus of elasticity $E$ is $523104\,kN/cm^2$. The objective function is therefore formulated as:

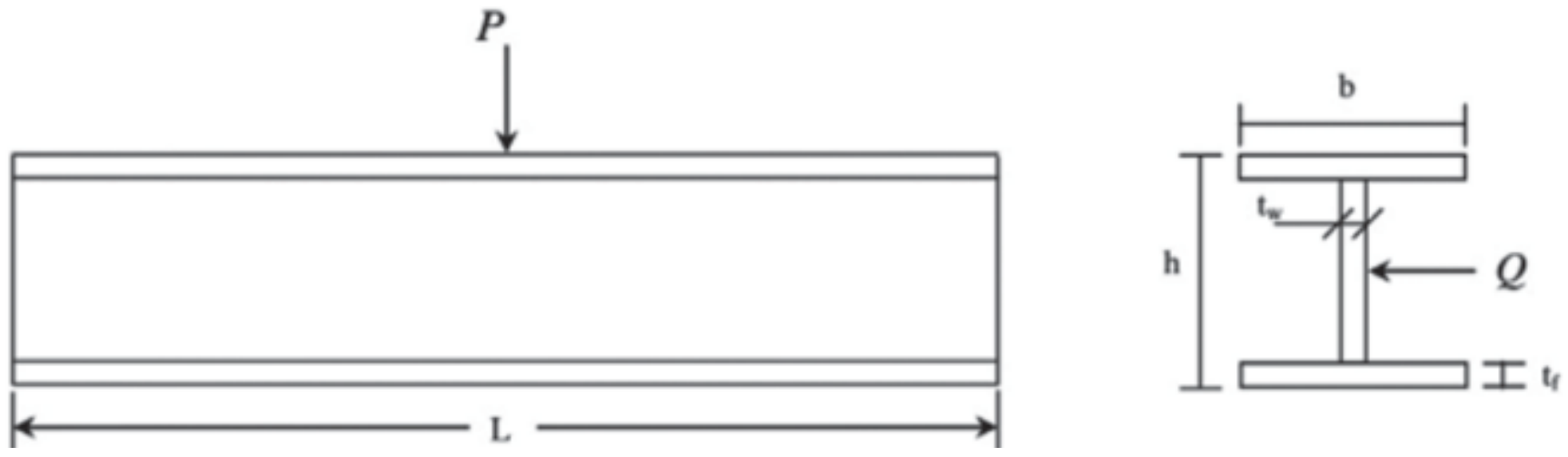

**Fig 15.** I-Beam design

$$Minimise: f\left(b, h, t_{w,} t_{f,}\right) = \frac{5000}{\frac{t_w(h-2t_f)}{12} + \frac{bt_f^3}{6} + 2bt_f\left(\frac{h-t_f}{2}\right)^2} \tag{7.2.25}$$

subject to a cross-section area of less than 300 cm$^2$

$$g_1 = 2bt_w + t_w\left(h - 2t_f\right) \leq 300 \tag{7.2.26}$$

If the maximum allowable bending stress for the beam is $56\,kN/cm^2$, the corresponding stress constraint is:

$$g_2 = \frac{18h \times 10^4}{t_w\left(h-2t_f\right)^3 + 2bt_w(4t_f^2 + 3h\left(h-2t_f\right))} + \frac{15b \times 10^3}{\left(h-2t_f\right)t_w^3 + 2b^3t_w} \leq 56 \tag{7.2.27}$$

$$10 \leq h \leq 80, \tag{7.2.28}$$

$$10 \leq b \leq 50, \tag{7.2.29}$$

$$0.9 \leq t_w \leq 5, \tag{7.2.30}$$

$$0.9 \leq t_f \leq 5. \tag{7.2.31}$$

**Table 13.** The Constant Terms Provided for the Formulation of the Helical Spring Design Problem

| Constant terms | Description | Values |
|---|---|---|
| $P_{max}$ | Maximum work load | $1000\ lb$ |
| $S$ | Maximum shear stress | $189e3psi$ |
| $E$ | Elastic module of material | $30e6psi$ |
| $G$ | Shear module of material | $11.5e6psi$ |
| $L_{free}$ | Maximum coil free length | $14.0in$ |
| $d_{min}$ | Minimum wire diameter | $0.2in$ |
| $D_{max}$ | Maximum outside diameter of spring | $3.0in$ |
| $P$ | Preload compression force | $300.0lb$ |
| $\delta_{pm}$ | Maximum deflection under preload | $6.0in$ |
| $\delta_w$ | Deflection from preload position to maximum load position | $1.25in$ |

In Table 14, the statistical performance of FA, GA, and FAGA are compared for solving the I-beam vertical deflection design problem. FAGA achieves the best deflection value which is identical to FA though superior to GA's. Furthermore, FAGA outperforms both algorithms in efficiency, as indicated by its average CPU time. FAGA also reduces computational effort with compared to GA.

**Table 14.** Statistical: I-beam Vertical deflection

| **Results** | **FA** | **GA** | **FAGA** |
|---|---|---|---|
| **Best** | 0.006625 | 0.006667 | 0.006625 |
| **Mean** | 0.007496 | 0.007532 | 0.006696 |
| **Worst** | 0.011725 | 0.010981 | 0.008423 |
| **Std. Dev** | 0.001334 | 0.001052 | 0.000329 |
| **Avg. CPU time (sec)** | 2.14 | 2.92 | 1.91 |
| **Avg. Fun. Eval.** | 2640 | 5108 | 4912 |

Table 15, highlights FAGA performance against other optimization algorithms, including MRSLS (Kale and Kulkarni, 2021), CS (Gandomi et al., 2013), SOS (Cheng and Prayoga, 2014), and CI-SAPF (Kale and Kulkarni, 2021). FAGA achieves the lowest deflection value of **0.6625e-2**, which shows similar performance to FA, CI-SAPF, MRSLS and GA. However, performs better than CS and SOS. FAGA further reduces the computational cost. This result highlights FAGA's robustness and its ability to deliver precise solutions with lower computational costs compared to traditional and hybrid optimization techniques.

**Table 15.** Performance comparison of various algorithms solving I-section beam vertical defection design problem

| **Techniques** | **MRSLS (Kale and Kulkarni, 2021)** | **CS (Gandomi et al., 2013)** | **SOS (Cheng and Prayoga, 2014)** | **CI-SAPF (Kale and Kulkarni, 2021)** | **FA** | **GA** | **FAGA** |
|---|---|---|---|---|---|---|---|
| $h$ | 80.0 | 80.0 | 80.0 | 80 | 80.0 | 79.7250 | 80.0 |
| $b$ | 50.0 | 50.0 | 50.0 | 50 | 50.0 | 49.9932 | 50.0 |
| $t_w$ | 1.7432 | 0.90 | 0.90 | 1.7647 | 1.7647 | 1.7620 | 1.7647 |
| $t_f$ | 4.9971 | 2.3216 | 2.3217 | 4.9999 | 5.0 | 4.9966 | 5.0 |
| **Deflection** $f(x)$ | 0.6645e-2 | 0.01307 | 0.01307 | 0.6626e-2 | 0.6626e-2 | 0.6667e-2 | 0.6625e-2 |
| **Function Evaluations** | 686 | 5000 | 5000 | 3900 | 1573 | 5168 | 4925 |

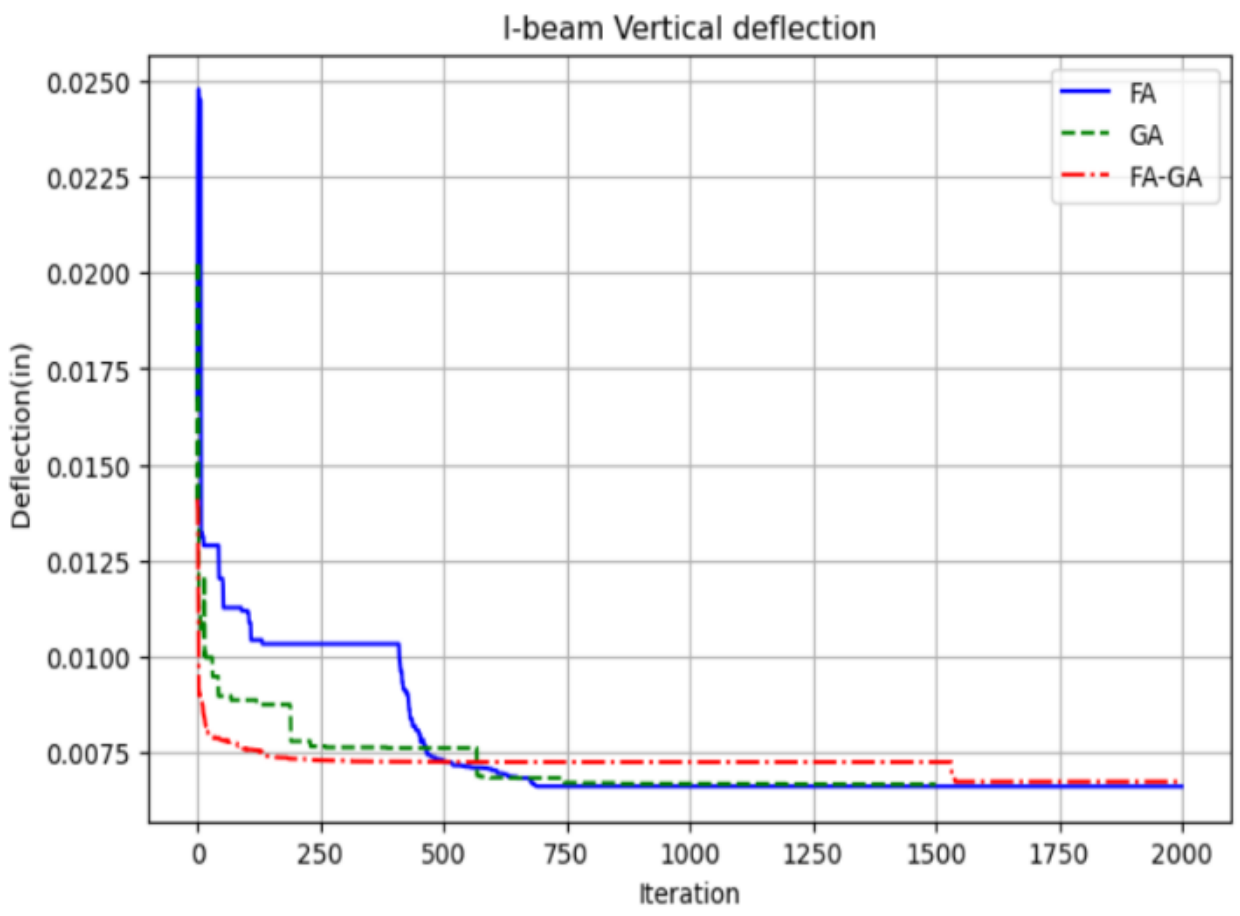

**Fig. 16.** Comparison of the Convergence Curves of FA, GA and FAGA for Solving I-section beam vertical defection design problem

The convergence curves for FA, GA, and FAGA for the I-beam vertical deflection problem are presented in Fig. 16. The graph shows that FAGA converges steadily compared to FA and GA. GA initially converges quickly, while FAGA maintains a steady decrease in deflection and achieves superior precision. While FA exhibits some exploration during the process, FAGA converges later in the process compared to both FA and GA, highlighting its ability to avoid getting trapped in local minima and deliver optimal solutions.

### 7.3.1 0-1 Single Knapsack Problem

The FAGA hybrid algorithm is validated by solving 0-1 knapsack problem, demonstrating its effectiveness on a variety of test cases. The test cases, named $f_1 - f_{20}$ , are derived from Kulkarni and Shabir (2016), which are well-known benchmark problems. These test cases are used to compare the performance of FAGA with other optimization algorithms, including Binary Kepler Optimization Algorithm (BKOA) (Abdel-Basset et al., 2023), Genetic Algorithm with Tournament selection (GAT) (Huang et al., 2006), Binary Marine Predators Algorithm (BMPA) (Beheshti and Zahra, 2022), standard Genetic Algorithm (GA) (Deb et al., 2002; Huang et al., 2006), Binary Jaya Algorithm (BJA) (Chaudhuri et al., 2022), and the Binary Young's Double-Slit Experiment (BYDSE) (Mohammed and Barakat, 2023) optimizer. Each algorithm brings a unique approach to solving knapsack problems, however has specific drawbacks. For instance, the BKOA may suffer from slow convergence rates in large search spaces, while BMPA sometimes be limited by premature convergence, affecting its ability to explore optimal solutions. GAT, while efficient for some combinatorial problems, may struggle with diversity issues, potentially leading to suboptimal results. GA, a widely used approach, also face similar limitations in maintaining diversity across generations. The FAGA hybrid algorithm combines the strengths of FA and GA, aiming to balance exploration and exploitation effectively. The proposed hybrid approach leverages FA attractiveness-driven movement mechanism to enhance the global search capabilities while using GA's crossover and mutation functions to ensure a more robust local search.

**Table 16**. Summary of SKPs solved using the FAGA hybrid algorithm

| Problem | Number of objects | Knapsack capacity | Solutions (f(v)) | | | Standard deviation | Total Time(s) | Average time(s) |
|---|---|---|---|---|---|---|---|---|
| | | | **Best** | **Mean** | **Worst** | | | |
| $f_1$ | 10 | 269 | 295 | 292.2 | 281 | 5.1621 | 0.93 | 0.031 |
| $f_2$ | 20 | 878 | 1024 | 1018.467 | 1009 | 5.3092 | 3.233 | 0.1077 |
| $f_3$ | 4 | 20 | 35 | 34.1333 | 33 | 1.008 | 0.3258 | 0.0108 |
| $f_4$ | 4 | 11 | 23 | 22.6897 | 22 | 0.4794 | 0.0987 | 0.0032 |
| $f_5$ | 15 | 375 | 481.0694 | 476.1648 | 426.0262 | 11.8268 | 1.2173 | 0.0405 |
| $f_6$ | 10 | 60 | 52 | 51.2333 | 50 | 0.6789 | 0.5052 | 0.0168 |
| $f_7$ | 7 | 50 | 107 | 102.1333 | 96 | 3.73 | 0.4632 | 0.0154 |
| $f_8$ | 23 | 10,000 | 9767 | 9758.233 | 9750 | 3.5689 | 4.7319 | 0.1577 |
| $f_9$ | 5 | 80 | 130 | 125.6 | 106 | 7.3794 | 0.4098 | 0.01366 |
| $f_{10}$ | 20 | 879 | 1025 | 1009.766 | 941 | 18.6153 | 2.7072 | 0.0902 |
| $f_{11}$ | 30 | 577 | 1437 | 1426.74 | 1409 | 9.3238 | 7.2115 | 0.2671 |
| $f_{12}$ | 35 | 655 | 1689 | 1682.2 | 1656 | 8.2276 | 11.929 | 0.3976 |
| $f_{13}$ | 40 | 819 | 1821 | 1813.833 | 1792 | 5.5651 | 23.187 | 0.7729 |
| $f_{14}$ | 45 | 907 | 2033 | 2018.933 | 2010 | 2.4344 | 24.774 | 0.8258 |
| $f_{15}$ | 50 | 882 | 2440 | 2437.467 | 2387 | 9.7582 | 46.636 | 1.5545 |
| $f_{16}$ | 55 | 1050 | 2651 | 2635.133 | 2613 | 8.6652 | 68.142 | 2.2714 |
| $f_{17}$ | 60 | 1006 | 2917 | 2915.533 | 2901 | 3.954337 | 153.69 | 5.1229 |
| $f_{18}$ | 65 | 1319 | 2818 | 2811.8 | 2802 | 4.8094 | 177.69 | 5.9229 |
| $f_{19}$ | 70 | 1426 | 3223 | 3217.967 | 3210 | 4.2384 | 261.93 | 8.73099 |
| $f_{20}$ | 75 | 1433 | 3614 | 3603 | 3591 | 7.7948 | 60.872 | 2.02907 |

Table 16 summarizes FAGA's performance across various knapsack test cases, measuring metrics such as best, mean, and worst fitness values, standard deviation, total time, and average time. The algorithm consistently achieves strong results, with minimal deviation in fitness values, indicating its reliability and robustness in finding optimal or near-optimal solutions within a manageable time. Notably, FAGA excels in larger test cases (e.g., $f_{15}$ to $f_{20}$), where both capacity and object counts are higher. In these scenarios, FAGA maintains a stable standard deviation and demonstrates efficient runtime, underscoring its capability to handle complex, large-scale problems without sacrificing accuracy or speed.

**Table 17**. Comparison of 0-1 Knapsack $f_1$ to $f_{20}$ problems obtained using FAGA with other algorithms

| Problem | Number of objects($N$) | Method | Optimal Solution $f(v)$ | Problem | Number of objects($N$) | Method | Optimal Solution $f(v)$ |
|---|---|---|---|---|---|---|---|
| $f_1$ | 10 | BKOA | 295 | $f_{11}$ | 30 | BKOA | 1437 |
| | | GAT | 295 | | | GAT | 1437 |
| | | BMPA | 295 | | | BMPA | 1431 |
| | | GA | 295 | | | GA | 1437 |
| | | BJA | 295 | | | BJA | 1437 |
| | | BYDSE | 295 | | | BYDSE | 1437 |
| | | FAGA | 295 | | | FAGA | 1437 |
| $f_2$ | 20 | BKOA | 1024 | $f_{12}$ | 35 | BKOA | 1689 |
| | | GAT | 1024 | | | GAT | 1689 |
| | | BMPA | 1024 | | | BMPA | 1689 |
| | | GA | 1024 | | | GA | 1689 |
| | | BJA | 1024 | | | BJA | 1689 |
| | | BYDSE | 1024 | | | BYDSE | 1586 |
| | | FAGA | 1024 | | | FAGA | 1689 |
| $f_3$ | 4 | BKOA | 35 | $f_{13}$ | 40 | BKOA | 1821 |
| | | GAT | 35 | | | GAT | 1821 |
| | | BMPA | 35 | | | BMPA | 1784 |
| | | GA | 35 | | | GA | 1821 |
| | | BJA | 35 | | | BJA | 1821 |
| | | BYDSE | 35 | | | BYDSE | 1701 |
| | | FAGA | 35 | | | FAGA | 1821 |
| $f_4$ | 4 | BKOA | 23 | $f_{14}$ | 45 | BKOA | 2033 |
| | | GAT | 23 | | | GAT | 2033 |
| | | BMPA | 23 | | | BMPA | 1985 |
| | | GA | 23 | | | GA | 2033 |
| | | BJA | 23 | | | BJA | 2033 |
| | | BYDSE | 23 | | | BYDSE | 1839 |
| | | FAGA | 23 | | | FAGA | 2033 |
| $f_5$ | 15 | BKOA | 481.069 | $f_{15}$ | 50 | BKOA | 2444 |
| | | GAT | 481.069 | | | GAT | 2444 |
| | | BMPA | 481.069 | | | BMPA | 2429 |
| | | GA | 481.069 | | | GA | 2444 |
| | | BJA | 481.069 | | | BJA | 2444 |
| | | BYDSE | 481.069 | | | BYDSE | 2206 |
| | | FAGA | 481.069 | | | FAGA | 2440 |
| $f_6$ | 10 | BKOA | 52 | $f_{16}$ | 55 | BKOA | 2651 |
| | | GAT | 52 | | | GAT | 2651 |
| | | BMPA | 52 | | | BMPA | 2593 |
| | | GA | 52 | | | GA | 2651 |
| | | BJA | 52 | | | BJA | 2651 |
| | | BYDSE | 52 | | | BYDSE | 2382 |
| | | FAGA | 52 | | | FAGA | 2651 |
| $f_7$ | 7 | BKOA | 107 | $f_{17}$ | 60 | BKOA | 2917 |
| | | GAT | 107 | | | GAT | 2917 |
| | | BMPA | 107 | | | BMPA | 2813 |
| | | GA | 107 | | | GA | 2917 |
| | | BJA | 107 | | | BJA | 2917 |
| | | BYDSE | 107 | | | BYDSE | 2544 |
| | | FAGA | 107 | | | FAGA | 2917 |
| $f_8$ | 23 | BKOA | 9767 | $f_{18}$ | 65 | BKOA | 2818 |
| | | GAT | 9767 | | | GAT | 2818 |
| | | BMPA | 9767 | | | BMPA | 2733 |
| | | GA | 9767 | | | GA | 2818 |
| | | BJA | 9751 | | | BJA | 2817 |
| | | BYDSE | 9767 | | | BYDSE | 2456 |
| | | FAGA | 9767 | | | FAGA | 2818 |
| $f_9$ | 5 | BKOA | 130 | $f_{19}$ | 70 | BKOA | 3221 |
| | | GAT | 130 | | | GAT | 3223 |
| | | BMPA | 130 | | | BMPA | 3135 |
| | | GA | 130 | | | GA | 3223 |
| | | BJA | 130 | | | BJA | 3223 |
| | | BYDSE | 130 | | | BYDSE | 2919 |
| | | FAGA | 130 | | | FAGA | 3223 |
| $f_{10}$ | 20 | BKOA | 1025 | $f_{20}$ | 75 | BKOA | 3609 |
| | | GAT | 1025 | | | GAT | 3614 |
| | | BMPA | 1025 | | | BMPA | 3397 |
| | | GA | 1025 | | | GA | 3614 |
| | | BJA | 1025 | | | BJA | 3609 |
| | | BYDSE | 1025 | | | BYDSE | 3039 |
| | | FAGA | 1025 | | | FAGA | 3614 |

In Table 17, the comparison highlights the best fitness values achieved by each algorithm across various 0-1 knapsack test cases, $f_1$ to $f_{20}$. This comparative analysis reveals that the FAGA hybrid algorithm consistently produces optimal solutions across multiple problem instances. For smaller instances, such as $f_1$ to $f_6$, all algorithms, including FAGA, attain the best fitness value, demonstrating similar performance due to the simplicity of these problems. However, in larger and more complex cases ($f_{10}$ onward), FAGA stands out by consistently achieving superior or comparable fitness values relative to many other optimization algorithms. FAGA delivers results comparable to BKOA and GA in most large cases, reaching the best or near-best fitness values consistently. In contrast, BMPA and tend to underperform as problem size increases, with BMPA frequently converging prematurely and BYDSE exhibiting greater variance in fitness outcomes. For instance, in cases such as $f_{13}$ , $f_{14}$ , and $f_{18}$ , FAGA surpasses BMPA and BYDSE by maintaining a higher and more stable fitness value. This comparison highlights FAGA's robustness and superior ability to balance exploration and exploitation, particularly in larger problem instances where other algorithms is likely to falter.

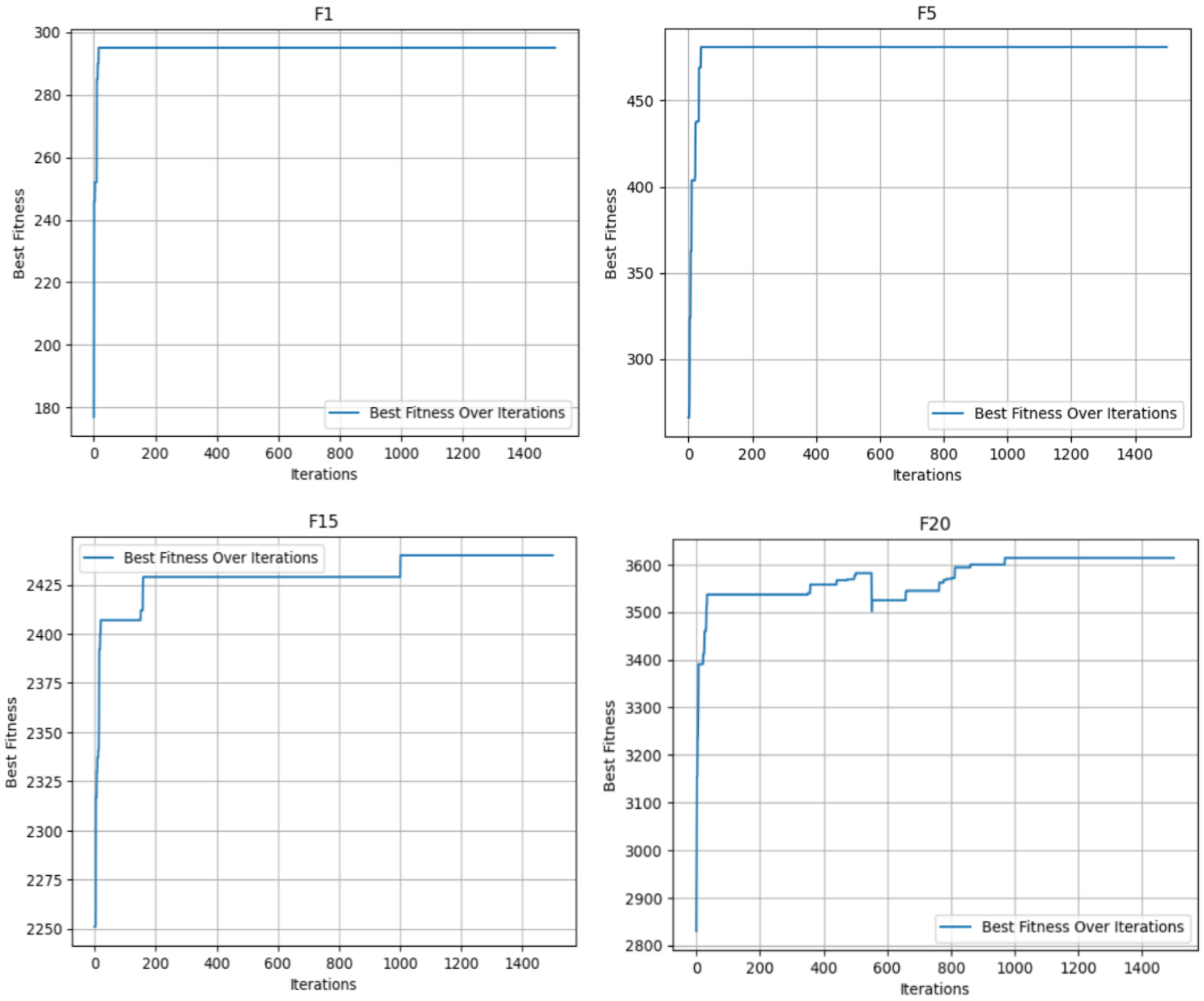

**Fig 17.** Convergence of each individual's values for different SKPs

The convergence behavior of FAGA for selected knapsack problems: $f_1$ , $f_5$ , $f_{15}$ , and $f_{20}$ is presented in Figure 17. These convergence graphs illustrate the FAGA's iterative optimization process, demonstrating how quickly the algorithm approaches the optimal solution over time. For smaller problems, such as $f_1$, FAGA attains the optimal solution within the first 50 iterations, indicating rapid convergence due to the relatively low complexity of the search space. In larger problems like $f_{15}$ and $f_{20}$, the convergence rate progressed more gradually yet remains consistent. Overall, the convergence behavior demonstrates FAGA's effectiveness as a reliable and scalable approach for solving diverse 0-1 knapsack problems.

## 7.3.2 0-1 Multidimensional Knapsack Problem

The performance of the hybrid FAGA algorithm in solving Multi-Knapsack Problems (MKP) is validated using benchmark datasets from the OR-Library (Beasley and John, 1990; Chih et al., 2014), specifically the "WEISH" dataset (Shih, 1979). This dataset consists of 30 problem instances commonly used to evaluate the effectiveness of optimization algorithms. The proposed algorithm is compared with six advanced binary optimization algorithms, namely Binary Modified Whale Optimization Algorithm (BIWOA) (Abdel-Basset et al., 2019), Binary Modified Multi-Verse Optimization (BMMVO) (Abdel-Basset et al., 2019), Binary Sin-Cosine Algorithm (BSCA) (Pinto et al., 2019), Binary Harris Hawks Algorithm (BHHA) (Heidari et al., 2019), Binary Squirrel Search Algorithm (BSSA) (Mirjalili et al., 2017), and Moth Search Algorithm (MS) (Wang, 2018; Li et al., 2022). Furthermore, FAGA's results are benchmarked against an earlier version of the FAGA (Nand and Sharma, 2019) algorithm to evaluate the improvements introduced by the modified version.

**Table 18.** Results after applying FAGA hybrid algorithm to the Weish dataset

| Problem Set | Number of objects | True Optimum | Solutions (f(v)) | | | Standard deviation | Avg. Function Evaluation | Total Time(s) | Average time(s) |
|---|---|---|---|---|---|---|---|---|---|
| | | | Best | Mean | Worst | | | | |
| Weish 1 | 30 | 4554 | 4554 | 4540.93 | 4513 | 13.45 | 7060 | 58.55 | 1.95 |
| Weish 2 | 30 | 4536 | 4536 | 4518.4 | 4500 | 16.33 | 6676 | 60.92 | 2.03 |
| Weish 3 | 30 | 4115 | 4115 | 4111.53 | 4106 | 3.92 | 735 | 8.68 | 0.29 |
| Weish 4 | 30 | 4561 | 4561 | 4546.5 | 4505 | 19.12 | 989 | 9.81 | 0.33 |
| Weish 5 | 30 | 4514 | 4514 | 4483.4 | 4451 | 29.29 | 6079 | 48.55 | 1.61 |
| Weish 6 | 40 | 5557 | 5557 | 5533.17 | 5500 | 22.29 | 5921 | 63.65 | 2.12 |
| Weish 7 | 40 | 5567 | 5567 | 5542.57 | 5447 | 26.09 | 6591 | 84.15 | 2.8 |
| Weish 8 | 40 | 5605 | 5605 | 5595.57 | 5542 | 15.66 | 9654 | 91.56 | 3.05 |
| Weish 9 | 40 | 5246 | 5246 | 5231.93 | 5200 | 19.04 | 7833 | 82.14 | 2.74 |
| Weish 10 | 50 | 6339 | 6339 | 6298.17 | 6202 | 41.36 | 4748 | 47.60 | 1.59 |
| Weish 11 | 50 | 5643 | 5643 | 5599.94 | 5512 | 53.42 | 33479 | 372.63 | 12.42 |
| Weish 12 | 50 | 6339 | 6339 | 6317.4 | 6250 | 28.41 | 9056 | 111.09 | 3.7 |
| Weish 13 | 50 | 6159 | 6159 | 6040.09 | 5775 | 108.41 | 11768 | 112.91 | 3.76 |
| Weish 14 | 60 | 6954 | 6954 | 6887.63 | 6804 | 49.77 | 19547 | 164.88 | 5.5 |
| Weish 15 | 60 | 7486 | 7486 | 7456.33 | 7385 | 31.95 | 8100 | 74.04 | 2.47 |
| Weish 16 | 60 | 7289 | 7289 | 7258.8 | 7145 | 40.81 | 8580 | 68.87 | 2.3 |
| Weish 17 | 60 | 8633 | 8633 | 8512.17 | 5670 | 28.37 | 5384 | 47.3 | 1.58 |
| Weish 18 | 70 | 9580 | 9580 | 9541.27 | 9458 | 37.66 | 26399 | 233.48 | 7.78 |
| Weish 19 | 70 | 7698 | 7698 | 7629.23 | 7545 | 57.74 | 21413 | 186.61 | 6.22 |
| Weish 20 | 70 | 9450 | 9450 | 9410.87 | 9352 | 38.78 | 29793 | 277.1 | 9.24 |
| Weish 21 | 70 | 9074 | 9074 | 9000.5 | 8945 | 37.5 | 25976 | 223.98 | 7.47 |
| Weish 22 | 80 | 8947 | 8947 | 8925.3 | 8854 | 23.42 | 46326 | 555.64 | 18.52 |
| Weish 23 | 80 | 8344 | 8344 | 8293.67 | 8251 | 31.52 | 42998 | 547.04 | 18.23 |
| Weish 24 | 80 | 10220 | 10220 | 10156 | 10120 | 22.36 | 46894 | 594.75 | 19.83 |
| Weish 25 | 80 | 9939 | 9939 | 9895.87 | 9853 | 26.7 | 24558 | 263.81 | 8.79 |
| Weish 26 | 90 | 9584 | 9584 | 9507.7 | 9451 | 34.19 | 69305 | 1513.8 | 28.52 |
| Weish 27 | 90 | 9819 | 9819 | 9770.4 | 9724 | 18.22 | 63573 | 1491.42 | 37.39 |
| Weish 28 | 90 | 9492 | 9492 | 9475.87 | 9447 | 14.66 | 59847.5 | 1452.58 | 42.83 |
| Weish 29 | 90 | 9410 | 9410 | 9389.13 | 9363 | 15.7 | 62029.93 | 1746.5 | 52.56 |
| Weish 30 | 90 | 11191 | 11191 | 11163.97 | 11135 | 14.86 | 60620 | 1738.27 | 20.14 |

In Table 18, the comprehensive summary of the results obtained by the FAGA algorithm for 30 MKP problems, evaluated over 30 trials each, is presented. The reported metrics include the best, mean, and worst fitness values, standard deviation, average function evaluations, total time, and average time. FAGA consistently demonstrates strong performance across these problems, with minimal variation between the best and worst fitness values. However, the standard deviation is notably high in certain instances, indicating variability in the results for specific problem sets. This observation suggests that while the algorithm performs well overall, solution quality fluctuates across trials for particular problem instances. Nonetheless, the best fitness values highlight FAGA's capability to achieve optimal or near-optimal results, especially in problems where the standard deviation remains lower. The average function evaluations and runtime fall within acceptable limits, demonstrating that FAGA efficiently tackle both small and large MKP instances, balancing solution quality and computational resources.

**Table 19.** Performance Comparison of Old and Modified FAGA Algorithms

| Problem | Optimal | Algorithm | Best fitness | Mean fitness | Problem | Optimal | Algorithm | Best fitness | Mean fitness |
|---|---|---|---|---|---|---|---|---|---|
| Weish01 | 4554 | FAGA (*2019)* | **4554** | 4545.96 | Weish16 | 7289 | FAGA (*2019)* | **7289** | 7253.68 |
| | | FAGA | **4554** | 4540.93 | | | FAGA | **7289** | 7258.8 |
| Weish02 | 4536 | FAGA (*2019)* | **4536** | 4534.12 | Weish17 | 8633 | FAGA (*2019)* | **8633** | 8626.24 |
| | | FAGA | **4536** | 4518.4 | | | FAGA | **8633** | 8512.17 |
| Weish03 | 4115 | FAGA (*2019)* | **4115** | 4106 | Weish18 | 9580 | FAGA (*2019)* | **9580** | 9556.2 |
| | | FAGA | **4115** | 4111.53 | | | FAGA | **9580** | 9541.27 |
| Weish04 | 4561 | FAGA (*2019)* | **4561** | 4558.76 | Weish19 | 7698 | FAGA (*2019)* | **7698** | 7580.24 |
| | | FAGA | **4561** | 4546.5 | | | FAGA | **7698** | 7629.23 |
| Weish05 | 4514 | FAGA (*2019)* | **4514** | 4506.44 | Weish20 | 9450 | FAGA (*2019)* | **9450** | 9400.12 |
| | | FAGA | **4514** | 4483.4 | | | FAGA | **9450** | 9410.87 |
| Weish06 | 5557 | FAGA (*2019)* | **5557** | 5549.28 | Weish21 | 9074 | FAGA (*2019)* | **9074** | 9034.08 |
| | | FAGA | **5557** | 5533.17 | | | FAGA | **9074** | 9000.5 |
| Weish07 | 5567 | FAGA (*2019)* | **5567** | 5545.64 | Weish22 | 8947 | FAGA (*2019)* | **8947** | 8856.72 |
| | | FAGA | **5567** | 5542.57 | | | FAGA | **8947** | 8925.3 |
| Weish08 | 5605 | FAGA (*2019)* | **5605** | 5594.20 | Weish23 | 8344 | FAGA (*2019)* | **8344** | 8203.71 |
| | | FAGA | **5605** | 5595.57 | | | FAGA | **8344** | 8293.67 |
| Weish09 | 5246 | FAGA (*2019)* | **5246** | 5215.76 | Weish24 | 10220 | FAGA (*2019)* | **10220** | 10204.92 |
| | | FAGA | **5246** | 5231.93 | | | FAGA | **10220** | 10156 |
| Weish10 | 6339 | FAGA (*2019)* | **6339** | 6310.24 | Weish25 | 9939 | FAGA (*2019)* | **9939** | 9889.32 |
| | | FAGA | **6339** | 6298.17 | | | FAGA | **9939** | 9895.87 |
| Weish11 | 5643 | FAGA (*2019)* | **5643** | 5571 | Weish26 | 9584 | FAGA (*2019)* | **9584** | 9502.18 |
| | | FAGA | **5643** | 5599.94 | | | FAGA | **9584** | 9507.7 |
| Weish12 | 6339 | FAGA (*2019)* | **6339** | 6301 | Weish27 | 9819 | FAGA (*2019)* | **9819** | 9683.1 |
| | | FAGA | **6339** | 6317.4 | | | FAGA | **9819** | 9770.4 |
| Weish13 | 6159 | FAGA (*2019)* | **6159** | 6121.84 | Weish28 | 9492 | FAGA (*2019)* | **9492** | 9163.18 |
| | | FAGA | **6159** | 6040.09 | | | FAGA | **9492** | 9475.87 |
| Weish14 | 6954 | FAGA (*2019)* | **6954** | 6904.36 | Weish29 | 9410 | FAGA (*2019)* | **9410** | 9262.6 |
| | | FAGA | **6954** | 6887.63 | | | FAGA | **9410** | 9389.13 |
| Weish15 | 7486 | FAGA (*2019)* | **7486** | 7442.54 | Weish30 | 11191 | FAGA (*2019)* | **11191** | 11169.72 |
| | | FAGA | **7486** | 7456.33 | | | FAGA | **11191** | 11163.97 |

The earlier version of the FAGA (Nand and Sharma, 2019) employs a step-by-step approach where the FA is applied during the first half of the iterations, and then the GA takes over for the second half. While this approach produces reasonable results, it exhibits drawbacks, including slower convergence and limited solution diversity. In contrast, the proposed FAGA integrates both algorithms throughout the entire optimization process, enabling them to collaborate simultaneously. This modification enhances the exploration of potential solutions and improves the exploitation of the best options identified. Consequently, the modified FAGA converges more rapidly to optimal solutions and delivers higher-quality results and reduces computational cost. The results presented in Table 19, indicate that both the earlier and proposed FAGA successfully identify the true optimal solution. However, in most of the problems the proposed algorithm achieves a mean fitness value that is closer to the optimal solution compared to the earlier version. This outcome demonstrates that the proposed FAGA outperforms the earlier approach and represents a more effective approach for optimization tasks.

**Table 20.** Comparison of results of the FAGA hybrid algorithm with other algorithms

| Problem | Knapsacks | Items | Algorithm | Optimal Solution $f(v)$ | Problem | Knapsacks | Items | Algorithm | Optimal Solution $f(v)$ |
|---|---|---|---|---|---|---|---|---|---|
| Weish01 | 5 | 30 | MS | 4554 | Weish16 | 5 | 60 | MS | 7289 |
| | | | BIWOA | 4554 | | | | BIWOA | 7289 |
| | | | BMMVO | 4554 | | | | BMMVO | 7289 |
| | | | BSCA | 4554 | | | | BSCA | 7289 |
| | | | BHHA | 4554 | | | | BHHA | 7289 |
| | | | BSSA | 4554 | | | | BSSA | 7289 |
| | | | FAGA | 4554 | | | | FAGA | 7289 |
| Weish02 | 5 | 30 | MS | 4536 | Weish17 | 5 | 60 | MS | 8633 |
| | | | BIWOA | 4536 | | | | BIWOA | 8633 |
| | | | BMMVO | 4536 | | | | BMMVO | 8624 |
| | | | BSCA | 4536 | | | | BSCA | 8633 |
| | | | BHHA | 4536 | | | | BHHA | 8633 |
| | | | BSSA | 4536 | | | | BSSA | 8633 |
| | | | FAGA | 4536 | | | | FAGA | 8633 |
| Weish03 | 5 | 30 | MS | 4106 | Weish18 | 5 | 70 | MS | 9540 |
| | | | BIWOA | 4106 | | | | BIWOA | 9560 |
| | | | BMMVO | 4106 | | | | BMMVO | 9456 |
| | | | BSCA | 4106 | | | | BSCA | 9573 |
| | | | BHHA | 4106 | | | | BHHA | 9580 |
| | | | BSSA | 4106 | | | | BSSA | 9573 |
| | | | FAGA | 4115 | | | | FAGA | 9580 |
| Weish04 | 5 | 30 | MS | 4561 | Weish19 | 5 | 70 | MS | 7698 |
| | | | BIWOA | 4561 | | | | BIWOA | 7698 |
| | | | BMMVO | 4561 | | | | BMMVO | 7698 |
| | | | BSCA | 4561 | | | | BSCA | 7698 |
| | | | BHHA | 4561 | | | | BHHA | 7698 |
| | | | BSSA | 4561 | | | | BSSA | 7698 |
| | | | FAGA | 4561 | | | | FAGA | 7698 |
| Weish05 | 5 | 30 | MS | 4514 | Weish20 | 5 | 70 | MS | 9450 |
| | | | BIWOA | 4514 | | | | BIWOA | 9450 |
| | | | BMMVO | 4514 | | | | BMMVO | 9445 |
| | | | BSCA | 4514 | | | | BSCA | 9450 |
| | | | BHHA | 4514 | | | | BHHA | 9450 |
| | | | BSSA | 4514 | | | | BSSA | 9450 |
| | | | FAGA | 4514 | | | | FAGA | 9450 |
| Weish06 | 5 | 40 | MS | 5557 | Weish21 | 5 | 70 | MS | 9074 |
| | | | BIWOA | 5557 | | | | BIWOA | 9074 |
| | | | BMMVO | 5557 | | | | BMMVO | 9074 |
| | | | BSCA | 5557 | | | | BSCA | 9074 |
| | | | BHHA | 5557 | | | | BHHA | 9074 |
| | | | BSSA | 5557 | | | | BSSA | 9074 |
| | | | FAGA | 5557 | | | | FAGA | 9074 |
| Weish07 | 5 | 40 | MS | 5567 | Weish22 | 5 | 80 | MS | 8790 |
| | | | BIWOA | 5567 | | | | BIWOA | 8909 |
| | | | BMMVO | 5567 | | | | BMMVO | 8886 |
| | | | BSCA | 5567 | | | | BSCA | 8909 |
| | | | BHHA | 5567 | | | | BHHA | 8912 |
| | | | BSSA | 5567 | | | | BSSA | 8912 |
| | | | FAGA | 5567 | | | | FAGA | 8947 |
| Weish08 | 5 | 40 | MS | 5605 | Weish23 | 5 | 80 | MS | 8170 |
| | | | BIWOA | 5605 | | | | BIWOA | 8303 |
| | | | BMMVO | 5605 | | | | BMMVO | 8250 |
| | | | BSCA | 5605 | | | | BSCA | 8344 |
| | | | BHHA | 5605 | | | | BHHA | 8344 |
| | | | BSSA | 5605 | | | | BSSA | 8344 |
| | | | FAGA | 5605 | | | | FAGA | 8344 |
| Weish09 | 5 | 40 | MS | 5246 | Weish24 | 5 | 80 | MS | 10,189 |
| | | | BIWOA | 5246 | | | | BIWOA | 10,189 |
| | | | BMMVO | 5246 | | | | BMMVO | 10,058 |
| | | | BSCA | 5246 | | | | BSCA | 10,215 |
| | | | BHHA | 5246 | | | | BHHA | 10,202 |
| | | | BSSA | 5246 | | | | BSSA | 10,220 |
| | | | FAGA | 5246 | | | | FAGA | 10220 |
| Weish10 | 5 | 50 | MS | 6339 | Weish25 | 5 | 80 | MS | 9922 |
| | | | BIWOA | 6323 | | | | BIWOA | 9885 |
| | | | BMMVO | 6303 | | | | BMMVO | 9844 |
| | | | BSCA | 6303 | | | | BSCA | 9939 |
| | | | BHHA | 6303 | | | | BHHA | 9939 |
| | | | BSSA | 6303 | | | | BSSA | 9939 |
| | | | FAGA | 6339 | | | | FAGA | 9939 |
| Weish11 | 5 | 50 | MS | 5643 | Weish26 | 5 | 90 | MS | 9581 |
| | | | BIWOA | 5643 | | | | BIWOA | 9575 |
| | | | BMMVO | 5643 | | | | BMMVO | 9575 |
| | | | BSCA | 5643 | | | | BSCA | 9575 |
| | | | BHHA | 5643 | | | | BHHA | 9575 |
| | | | BSSA | 5643 | | | | BSSA | 9575 |
| | | | FAGA | 5643 | | | | FAGA | 9584 |
| Weish12 | 5 | 50 | MS | 6339 | Weish27 | 5 | 90 | MS | 9764 |
| | | | BIWOA | 6302 | | | | BIWOA | 9778 |
| | | | BMMVO | 6301 | | | | BMMVO | 9589 |
| | | | BSCA | 6302 | | | | BSCA | 9764 |
| | | | BHHA | 6302 | | | | BHHA | 9764 |
| | | | BSSA | 6302 | | | | BSSA | 9764 |
| | | | FAGA | 6339 | | | | FAGA | 9819 |
| Weish13 | 5 | 50 | MS | 6159 | Weish28 | 5 | 90 | MS | 9492 |
| | | | BIWOA | 6159 | | | | BIWOA | 9454 |
| | | | BMMVO | 6159 | | | | BMMVO | 9400 |
| | | | BSCA | 6159 | | | | BSCA | 9454 |
| | | | BHHA | 6159 | | | | BHHA | 9454 |
| | | | BSSA | 6159 | | | | BSSA | 9454 |
| | | | FAGA | 6159 | | | | FAGA | 9492 |
| Weish14 | 5 | 60 | MS | 6954 | Weish29 | 5 | 90 | MS | 9369 |
| | | | BIWOA | 6923 | | | | BIWOA | 9369 |
| | | | BMMVO | 6923 | | | | BMMVO | 9369 |
| | | | BSCA | 6923 | | | | BSCA | 9369 |
| | | | BHHA | 6923 | | | | BHHA | 9369 |
| | | | BSSA | 6923 | | | | BSSA | 9369 |
| | | | FAGA | 6954 | | | | FAGA | 9410 |
| Weish15 | 5 | 60 | MS | 7486 | Weish30 | 5 | 90 | MS | 11,148 |
| | | | BIWOA | 7486 | | | | BIWOA | 11,121 |
| | | | BMMVO | 7486 | | | | BMMVO | 11,025 |
| | | | BSCA | 7486 | | | | BSCA | 11,169 |
| | | | BHHA | 7486 | | | | BHHA | 11,169 |
| | | | BSSA | 7486 | | | | BSSA | 11,169 |
| | | | FAGA | 7486 | | | | FAGA | 11,191 |

In Table 20, the comparison between the proposed FAGA and other state-of-the-art algorithms are presented, measured against known optimal solutions for each MKP instance. The data reveals that FAGA performs competitively, matching or surpassing the performance of algorithms such as BIWOA, BSSA and BHHA on several complex instances. While some algorithms, like BIWOA and BMMVO, exhibit strong performance on simpler problems, FAGA demonstrates superior results on more challenging instances, finding solutions closer to the optimal. FAGA's balance of exploration and exploitation enables it to perform consistently across all problem types. In many cases, FAGA achieves results similar to the optimal solutions, outperforming most other algorithms in both solution quality and computational efficiency after 30 trials.

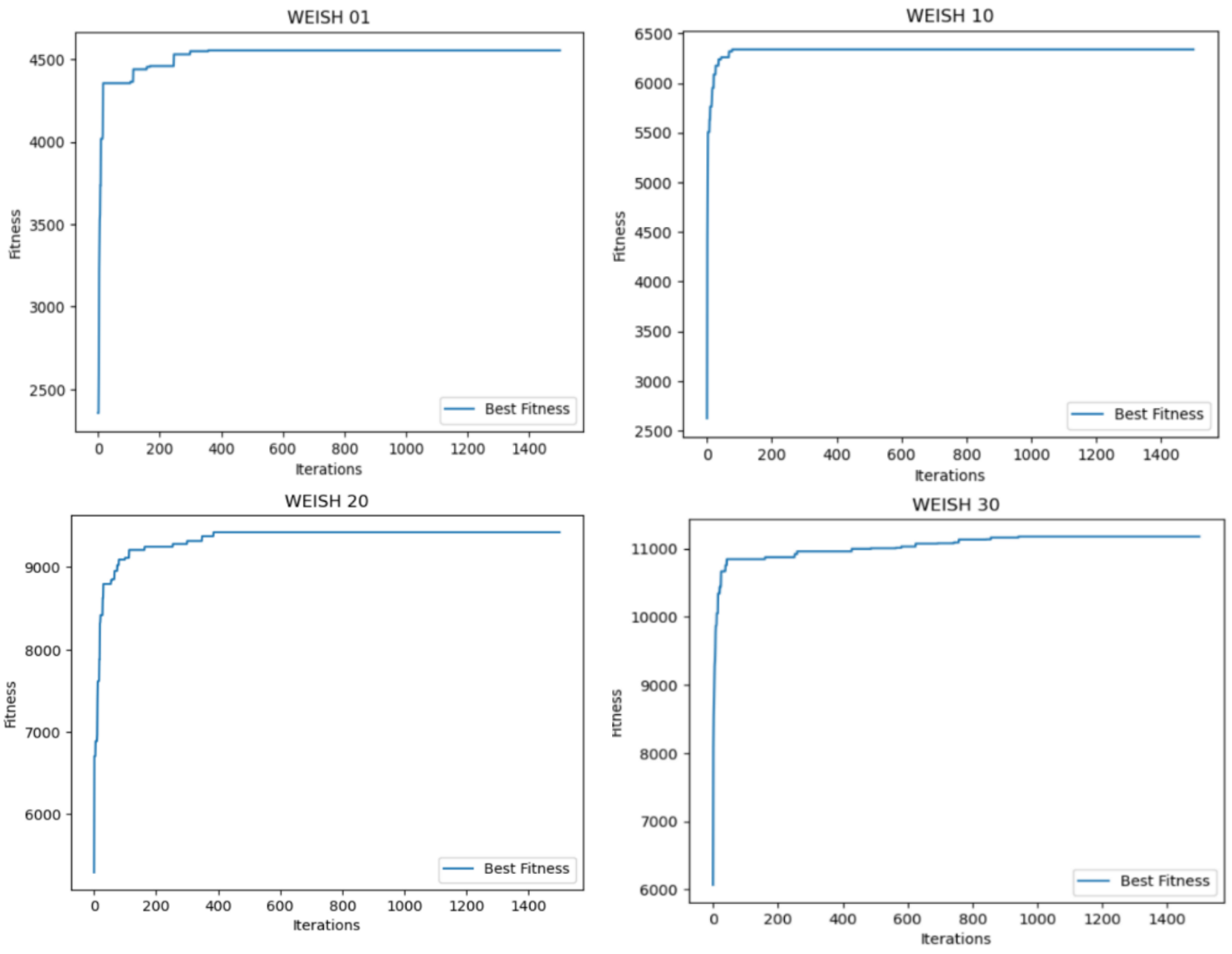

**Fig 18.** Convergence of individual's values for different MKPs

The convergence graphs in Fig 18, for the FAGA algorithm on four instances from the WEISH dataset (WEISH 01, WEISH 10, WEISH 20, and WEISH 30) illustrate its performance in solving varying levels of complexity in MKP's. For WEISH 01 and WEISH 10, FAGA achieves rapid convergence, indicating that it efficiently reaches near-optimal solutions for smaller or less complex problems. In more complex MKP problem, such as WEISH 30, FAGA exhibits a gradual increase in fitness, converging at a slower pace while steadily progressing toward optimal values. This gradual improvement highlights FAGA's balanced exploration and exploitation capabilities, which become more evident in challenging problem sets. Overall, these results demonstrate FAGA's robustness and versatility, with consistent convergence behavior across problem complexities and minimal performance fluctuation, indicating reliable results across trials.

## Conclusion

The proposed hybrid FAGA algorithm demonstrates exceptional effectiveness in solving a wide range of optimization problems, undergoing rigorous testing using benchmark functions to evaluate its performance across various problem types, including unimodal and multimodal optimization functions. Results consistently show that the hybrid approach outperforms standalone FA and GA, achieving faster convergence and superior-quality solutions. When applied to real-world engineering design problems, FAGA delivers optimal results, surpassing existing methods in terms of solution quality and computational efficiency. By adhering to complex constraints and minimizing variations, the algorithm highlights its robustness and reliability for engineering applications.

FAGA performance in the combinatorial 0-1 Knapsack Problem showcases its capability to deliver optimal solutions for both single-constraint and multi-constraint cases. With lower standard deviations and means closer to the true optimum, the algorithm exhibits reduced variation in finding the best solutions. It consistently outperforms other algorithms for single and multi-constraint problems while demonstrating significant over earlier and existing approaches. Future advancements, such as machine learning-based parameter adaptation and parallel computation techniques, will further accelerate the optimization process and enable FAGA to address high-dimensional, multi-objective, and large-scale problems with enhanced accuracy and efficiency. These developments establish FAGA as a powerful and versatile approach for both theoretical and practical optimization challenges.

## Appendix

| $f$ | Number of objects ($N$) | Parameters | | |
|---|---|---|---|---|
| | | W | w | v |
| $f_1$ | 10 | 269 | 95, 4, 60, 32, 23, 72, 80, 62, 65, 46 | 55, 10, 47, 5, 4, 50, 8, 61, 85, 87 |
| $f_2$ | 20 | 878 | 92, 4, 83, 43, 88, 64, 98, 82, 6, 44, 32, 18, 56, 23, 85, 96, 70, 48, 14, 58 | 44, 46, 90, 72, 91, 40, 75, 35, 8, 54, 78, 40, 77, 15, 61, 17, 75, 29, 75, 63 |
| $f_3$ | 4 | 20 | 6, 5, 9, 7 | 9, 11, 13, 15 |
| $f_4$ | 4 | 11 | 2, 4, 6, 7 | 6, 10, 12, 13 |
| $f_5$ | 15 | 375 | 56.358531, 80.874050, 47.987304, 89.596240, 74.660482, 85.894345, 51.353496, 1.4989459, 36.445204, 16.5898624, 44.569231, 0.466933, 37.788018, 57.118442, 60.716575 | 0.125126, 19.330424, 58.500931, 35.029145, 82.284005, 17.410810, 71.050142, 30.399487, 9.140294, 14.731285, 98.852504, 11.908322, 0.891140, 53.166295, 60.716397 |
| $f_6$ | 10 | 60 | 30, 25, 20, 18, 17, 11, 5, 2, 1, 1 | 20, 18, 17, 15, 15, 10, 5, 3, 1, 1 |
| $f_7$ | 7 | 50 | 31, 10, 20, 19, 4, 3, 6 | 70, 20, 39, 37, 7, 5, 10 |
| $f_8$ | 23 | 10000 | 983, 982, 981, 980, 979, 978, 488, 976, 972, 486, 486, 972, 485, 969, 966, 483, 964, 963, 961, 958, 959 | 981,980,979,978,977,976,487, 974,970, 485, 485, 970, 484, 484, 976, 974, 962, 961, 959, 958, 857 |
| $f_9$ | 5 | 80 | 15, 20, 17, 8, 31 | 33, 24, 36, 37, 12 |
| $f_{10}$ | 20 | 879 | 84, 83, 43, 4, 44, 6, 82, 92, 23, 58, 16, 58, 14, 48, 70, 96, 32, 68, 92 | 91, 72, 90, 46, 55, 8, 35, 75, 61, 75, 17, 78, 40, 44, 77, 63, 75, 29, 75, 63 |
| $f_{11}$ | 30 | 577 | 46, 17, 35, 1, 26, 17, 17, 48, 38, 17, 32, 21, 29, 48, 31, 8, 42, 37, 6, 9, 15, 22, 27, 14, 42, 40, 14, 31, 6, 34 | 57, 64, 50, 6, 52, 6, 85, 60, 70, 65, 63, 96, 18, 48, 85, 50, 77, 18, 70, 92, 17, 43, 5, 23,67,88,35,3,91,48 |
| $f_{12}$ | 35 | 655 | 7, 4, 46, 47, 6, 33, 8, 35, 32, 3, 40, 50, 22, 18, 3, 12, 30, 31, 13, 33, 4, 48, 5, 17, 33, 26, 27, 19, 39, 15, 33, 47, 17, 41, 40 | 35, 67, 30, 69, 40, 40, 21, 73, 82, 93, 52, 20, 61, 20, 42, 86, 43, 93, 38, 70, 59, 11, 42, 93, 6, 39, 25, 23, 36, 93, 51, 81, 36, 46, 96 |
| $f_{13}$ | 40 | 819 | 28, 23, 35, 38, 30, 29, 11, 48, 26, 14, 12, 48, 35, 36, 33, 39, 30, 46, 22, 21, 10, 15, 46, 43, 19, 32, 2, 47, 24, 26, 39, 17, 32, 17, 16, 33, 22, 6, 12 | 13, 16, 42, 69, 66, 68, 1, 13, 77, 85, 75, 95, 92, 23, 51, 79, 53, 62, 56, 74, 7, 50, 23, 34, 56, 75, 42, 51, 13, 22, 30, 45, 25, 27, 90, 59, 94, 62, 26, 11 |
| $f_{14}$ | 45 | 907 | 18, 12, 38, 12, 23, 13, 18, 46, 1, 7, 40, 23, 11, 47, 49, 19, 50, 19, 7, 33, 4, 31, 35, 41, 42, 2, 33, 14, 48, 40, 12, 35, 17, 38, 50, 14, 47, 35, 5, 41, 24, 45, 39, 1 | 70, 78, 06, 33, 2, 58, 4, 27, 40, 45, 77, 63, 32, 30, 8, 18, 73, 92, 43, 38, 50, 78, 16, 38, 0, 40, 43, 43, 22, 50, 4, 57, 5, 88, 87, 34, 98, 96, 99, 16, 1, 25 |
| $f_{15}$ | 50 | 882 | 15, 40, 22, 28, 50, 35, 49, 5, 45, 3, 7, 32, 19, 16, 40, 16, 31, 24, 15, 42, 29, 4, 14, 9, 29, 11, 25, 37, 48, 39, 5, 47, 49, 31, 48, 17, 46, 1, 25, 8, 16, 9, 30, 33, 18, 3, 3, 4, 1 | 78, 69, 87, 59, 63, 12, 22, 4, 45, 33, 29, 50, 19, 94, 95, 60, 1, 91, 69, 8, 100, 32, 81, 47, 59, 48, 56, 18, 59, 16, 45, 54, 47, 84, 100, 98, 75, 20, 4, 19, 58, 63, 37, 64, 90, 26, 29, 13, 53, 83 |
| $f_{16}$ | 55 | 1050 | 27, 15, 46, 5, 40, 9, 36, 12, 11, 11, 49, 20, 32, 3, 12, 44, 24, 1, 24, 42, 44, 16, 12, 42, 22, 60, 10, 8, 46, 50, 20, 42, 48, 45, 43, 35, 9, 12, 22, 2, 14, 50, 16, 29, 31, 40, 35, 11, 4, 32, 35, 15, 29, 16 | 98, 74, 76, 4, 12, 27, 90, 98, 100, 30, 93, 19, 75, 72, 66, 83, 79, 78, 79, 44, 35, 6, 82, 11, 1, 28, 95, 68, 39, 86, 68, 61, 44, 97, 83, 2, 15, 49, 59, 30, 44, 40, 14, 96, 37, 84, 5, 43, 8, 32, 95, 86, 18 |
| $f_{17}$ | 60 | 1006 | 7, 13, 47, 33, 38, 41, 3, 21, 37, 7, 32, 13, 42, 42, 23, 49, 1, 20, 25, 31, 4, 8, 33, 11, 6, 3, 9, 26, 44, 39, 7, 4, 34, 25, 25, 16, 47, 46, 23, 38, | 81, 37, 70, 64, 97, 21, 60, 9, 55, 85, 5, 33, 71, 87, 51, 100, 43, 27, 48, 17, 26, 17, 76, 61, 97, 78, 58, 46, 29, 76, 10, 11, 74, 56, 39, 50, 72, 37, |
| | | | 10, 5, 11, 28, 34, 47, 3, 9, 22, 24, 41, 45, 10, 29, 1, 33, 16, 14 | 72, 100, 9, 47, 10, 73, 92, 9, 52, 56, 69, 30, 61, 26, 70, 46, 14, 27,9,3 |
| $f_{18}$ | 65 | 1319 | 4, 23, 48, 14, 35, 33, 11, 10, 40, 32, 23, 45, 9, 41, 47, 3, 26, 38, 2, 17, 19, 14, 32, 48, 34, 17, 50, 32, 38, 35, 18, 43, 19, 1, 24, 46, 9, 47, 38, 43, 23, 12, 30, 47, 17, 50, 43, 11, 3, 10, 7, 6, 30, 13, 48, 16, 47, 9, 24, 33, 36, 15, 47, 7, 14, 39 | 84, 65, 44, 61, 2, 48, 30, 64, 73, 80, 32, 47, 93, 15, 77, 69, 98, 14, 70, 18, 28, 97, 65, 77, 1, 85, 27, 95, 21, 14, 64, 60, 67, 42, 85, 85, 47, 19, 28, 4, 28, 50, 29, 70, 71, 94, 49, 44, 3, 8, 82, 97, 35, 43, 24, 37, 78, 71, 26, 66, 82, 93, 47, 92, 89 |
| $f_{19}$ | 70 | 1426 | 4, 16, 16, 2, 9, 44, 33, 43, 14, 45, 11, 49, 21, 12, 41, 19, 26, 38, 42, 20, 5, 14, 40, 47, 29, 47, 30, 50, 39, 10, 26, 34, 44, 31, 50, 7, 15, 24, 7, 12, 10, 34, 17, 40, 28, 12, 35, 3, 29, 20, 19, 9, 44, 14, 43, 41, 10, 49, 39, 31, 25, 46, 6, 7, 43 | 66, 76, 71, 61, 7, 30, 34, 65, 22, 8, 99, 21, 99, 62, 25, 72, 26, 12, 55, 22, 32, 98, 31, 95, 42, 12, 16, 100, 66, 45, 27, 19, 11, 83, 43, 93, 53, 88, 36, 41, 60, 92, 16, 14, 40, 92, 30, 58, 79, 33, 70, 35, 41, 84, 21, 30, 54, 63, 28, 61, 85, 71, 40, 58, 25, 73, 35 |
| $f_{20}$ | 75 | 1433 | 24, 45, 15, 40, 9, 37, 13, 5, 43, 35, 48, 50, 27, 46, 24, 45, 2, 7, 38, 40, 27, 15, 20, 5, 47, 21, 22, 33, 11, 45, 24, 37, 31, 46, 12, 12, 14, 41, 36, 44, 36, 34, 22, 29, 50, 18, 21, 28, 4, 20, 44, 45, 25, 11, 35, 24, 9, 40, 45, 8, 47, 12, 2, 1, 12, 36, 35, 14, 17, 5 | 2, 73, 82, 12, 49, 35, 78, 29, 83, 18, 87, 93, 20, 6, 55, 1, 83, 91, 47, 35, 51, 59, 94, 90, 81, 80, 84, 7, 51, 3, 17, 18, 38, 75, 73, 29, 24, 14, 29, 44, 41, 100, 37, 67, 82, 30, 39, 30, 91, 50, 21, 3, 18, 31, 97, 79, 68, 85, 43, 71, 49, 83, 44, 46,1, 100, 28, 4, 16 |